\documentclass[runningheads]{llncs}

\usepackage{eccv}

\usepackage{eccvabbrv}

\usepackage{graphicx}
\usepackage{booktabs}

\usepackage[accsupp]{axessibility}  

\usepackage{hyperref}

\usepackage{orcidlink}

\usepackage{multirow}
\usepackage{float}
\usepackage{caption}
\usepackage{comment}
\usepackage{wrapfig}

\usepackage{adjustbox}
\usepackage{makecell}
\usepackage{tabularx}

\begin{document}

\title{CustomX: Unified Character, Action, and Scene Customization in Video World Models}

\titlerunning{CustomX}

\author{Yitong Wang\inst{1,2}\textsuperscript{$\star$}\orcidlink{0009-0006-1373-4427} \and Fangyun Wei\inst{2}\textsuperscript{$\star$}\orcidlink{0000-0001-8784-4916} \and Hongyang Zhang\inst{3}\orcidlink{0000-0002-0548-6068} \and \\Bo Dai\inst{4}\textsuperscript{$\dagger$}\orcidlink{0000-0003-0777-9232} \and Yan Lu\inst{2}\orcidlink{0000-0001-5383-6424}}

\authorrunning{Y.~Wang et al.}


\institute{\textsuperscript{1}Fudan University\qquad \textsuperscript{2}Microsoft Research\\ \textsuperscript{3}University of Waterloo\qquad \textsuperscript{4}The University of Hong Kong\\{\small \texttt{wangyitong23@m.fudan.edu.cn} \quad \texttt{\{fawe, yanlu\}@microsoft.com}}\\{\small \texttt{hongyang.zhang@uwaterloo.ca} \quad \texttt{bdai@hku.hk}}\\{\small \url{https://snowflakewang.github.io/CustomX_Page/}}}

\maketitle

\renewcommand{\thefootnote}{}%
\footnotetext{$\star$~Equal contribution.\quad $\dagger$~Corresponding author.} \renewcommand{\thefootnote}{\arabic{footnote}}%

 \begin{center}
     \centering
     \captionsetup{type=figure}
     \includegraphics[width=\textwidth]{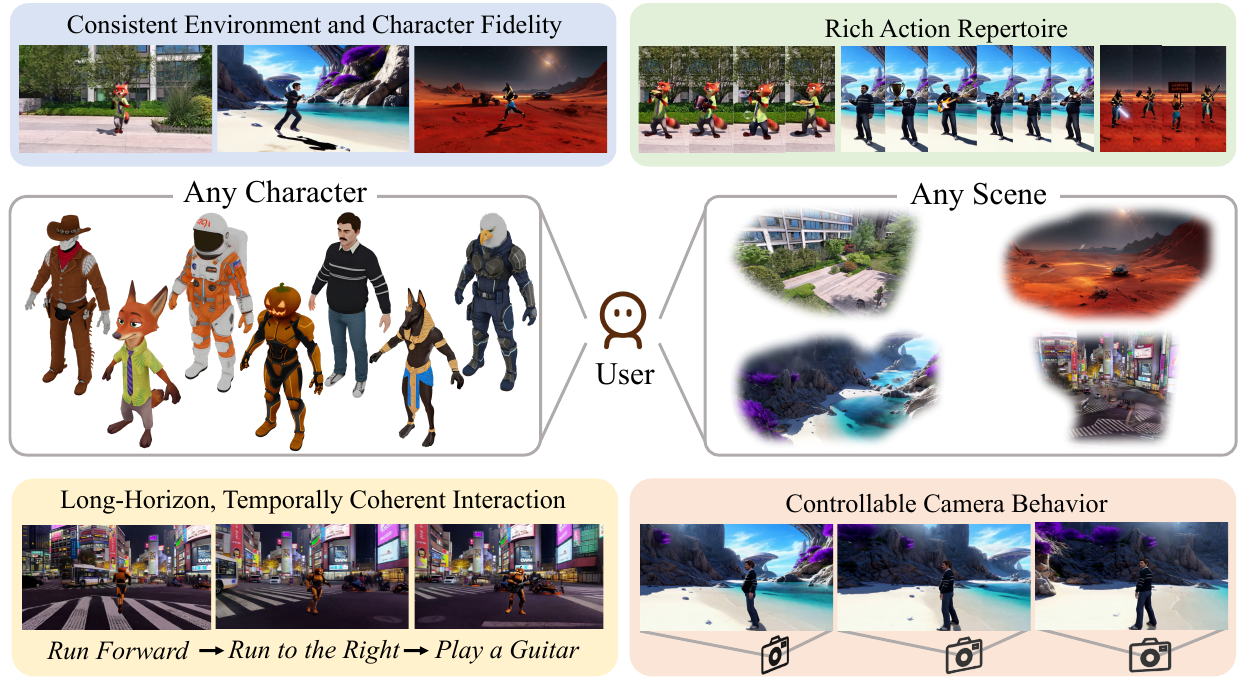}
    \captionof{figure}{\textbf{CustomX} enables users to provide a 3D or multi-view character together with a 3DGS scene, supporting iterative, multi-condition control of character behaviors and active environment exploration through natural language–guided video generation. The system features: (1) \textit{Consistent Environment and Character Fidelity}, ensuring visual and spatial coherence with the user-provided scene and character; (2) a \textit{Rich Action Repertoire} covering a wide range of behaviors, including locomotion, gestures, and object-centric interactions; (3) \textit{Long-Horizon, Temporally Coherent Interaction}, enabling iterative user interaction while maintaining continuity across generated clips; and (4) \textit{Controllable Camera Behavior}, which explicitly incorporates camera control, analogous to navigating 3DGS views, to produce accurate, user-specified viewpoints.}
    \label{fig:teaser}
 \end{center}

\begin{abstract}
Recent advances in world models have greatly enhanced interactive environment simulation. Existing methods mainly fall into two categories: (1) static world generation models, which construct 3D environments without active agents, and (2) controllable-entity models, which allow a single entity to perform limited actions in an otherwise uncontrollable environment. In this work, we introduce CustomX, leveraging the realism and structural grounding of static world generation while extending controllable-entity models to support user-specified characters capable of performing open-ended actions. Users can provide a 3DGS scene and a character, then use natural language to direct the character to perform diverse behaviors, ranging from basic locomotion to object-centric interactions, while freely exploring the environment. CustomX synthesizes temporally coherent video clips that preserve visual fidelity with the provided scene and character, formulated as a conditional autoregressive video generation problem. Built upon a pre-trained video generator, our training strategy significantly enhances motion dynamics while maintaining generalization across actions and characters. Our evaluation covers a broad range of aspects, including visual quality, character consistency, action controllability, and long-horizon coherence.
\end{abstract}    
\section{Introduction}
\label{sec:intro}
Recent advances in world models have led to substantial progress in simulating dynamic and interactive environments. Existing methods generally fall into two categories: (1) static world generation approaches~\cite{yang2025matrix3d, liu2025worldmirror, worldlabs}, which construct explorable 3D environments but lack active agents; and (2) controllable-entity approaches~\cite{li2025hunyuangamecraft, ye2025yan}, which allow a single agent to execute only a limited set of actions, such as steering a vehicle along a predefined path~\cite{feng2024thematrix, sun2025realplay, he2025matrix}, while leaving the environment itself uncontrollable. In this work, we propose an alternative framework that combines the strengths of both paradigms: leveraging the realism and structural grounding of static world generation while extending controllable-entity models to support user-specified characters capable of performing open-ended actions.

Specifically, users can provide a 3D Gaussian Splatting (3DGS) scene~\cite{kerbl3Dgaussians}, which represents either a synthetic environment or a real-world reconstruction, along with a 3D or multi-view character. Through natural language instructions, users can control the character’s behavior and enable active exploration within the environment. At each iteration, the model generates a video clip that captures the evolving states of both the character and the environment, resulting in coherent and temporally consistent generation. We name our method \textit{CustomX}. As illustrated in \cref{fig:teaser}, CustomX exhibits several key capabilities:

\begin{enumerate}
\item \textbf{Consistent Environment and Character Fidelity.} The visual contents appearing in the generated video clips exhibit strong consistency in visual identity and spatial layout with the user-provided scene and character.

    \item \textbf{Rich Action Repertoire.} Unlike previous works~\cite{li2025hunyuangamecraft, he2025matrix, chen2025deepverse} that limit the controllable entity to basic locomotion, our model enables the character to perform up to hundreds of distinct actions, encompassing not only navigation behaviors (e.g., ``moving forward'', ``turning left'') but also body-language gestures (e.g., ``waving hands'', ``saluting'') and object-centric interactions (e.g., ``making a phone call'', ``playing a guitar'').
    \item \textbf{Text Instruction as the Interface.} Users can directly guide the character through natural language commands.
    \item \textbf{Long-Horizon, Temporally Coherent Interaction.} Users can interact with the model iteratively, generating new video clips that remain temporally consistent with previously produced sequences.
    \item \textbf{Controllable Camera Behavior.} Our model supports flexible and intuitive camera control, allowing behaviors such as following a character’s trajectory or orbiting around it to achieve user-specified viewpoints. Unlike previous methods~\cite{he2024cameractrl, he2025cameractrl2} that encode camera trajectories into Plücker embeddings~\cite{sitzmann2021light} and inject them as conditioning signals into the generation network, our approach achieves camera control in a more geometrically grounded manner. Specifically, given a user-provided 3DGS scene and a defined camera path, we directly render a projection scene video along the specified trajectory. This rendered video serves as an explicit conditioning input, enabling the generation model to produce videos that accurately follow the desired camera motion.
\end{enumerate}

We formulate the entire process as a conditional autoregressive video generation problem. Concretely, the objective is to synthesize a video clip at each iteration, conditioned on a set of multi-modal inputs including: (a) the user-provided scene and character, which establish the spatial and visual grounding; (b) a text instruction, which specifies the intended behavior of the character; and (c) the previously generated video clips, which serve as temporal references to ensure consistency across iterations. 

Additionally, we adopt a pre-trained video generator~\cite{hu2025hunyuancustom} as the foundation of our framework. We find that fine-tuning it on a small dataset containing basic locomotion actions across diverse characters not only preserves the generalization ability of the pre-trained model but also enhances overall motion quality compared to the original generator. This phenomenon is analyzed in \cref{sec:main}.

In our experiments, we comprehensively evaluate the proposed model from multiple perspectives, including: (a) visual quality, assessed using the WorldScore benchmark~\cite{duan2025worldscore}; (b) character consistency, which measures the alignment between the character appearing in the generated video and the user-provided reference; (c) action control success rate, which quantifies how accurately the character’s behavior follows the input text instruction across a diverse set of up to around 150 actions; and (d) long-horizon generation quality, which evaluates the model’s ability to maintain temporal coherence and visual fidelity over extended interaction sequences. We compare CustomX with both video generation foundation models~\cite{wan2025,kong2024hunyuanvideo, yang2024cogvideox} and dedicated world models~\cite{li2025hunyuangamecraft,he2025matrix, chen2025deepverse}. Experimental results show that our method consistently outperforms both categories across nearly all evaluation metrics.
\section{Related Works}


\noindent\textbf{Controllable Video Generation.}
Recent foundation models for video generation~\cite{blattmann2023svd, yang2024cogvideox, HaCohen2024LTXVideo, kong2024hunyuanvideo, wan2025, veo, kling, seedance} have greatly improved modality alignment across text, image, and video, enabling large-scale pre-training for both text-to-video and image-to-video synthesis. Building on these advances, subsequent research has pursued finer-grained controllability by introducing mechanisms such as explicit subject control~\cite{jiang2024videobooth, vace, liu2025phantomvideo, fei2025skyreelsa2, chen2025alchemist, ju2025fulldit, liang2025omniv2v, chen2025omniinsert} and camera control~\cite{wang2024motionctrl, guo2023animatediff, yang2024direct, feng2024i2vcontrol, zhou2025seva, yu2024viewcrafter, yu2025trajectorycrafter, gu2025das, ren2025gen3c, yang2025omnicam, bai2025recammaster, luo2025camclonemaster}. For subject control, a typical approach~\cite{hu2025hunyuancustom, li2025bindweave} is to extract visual embeddings from reference images and use them within a Multimodal Diffusion Transformer~\cite{peebles2023dit, esser2024mmdit} to guide the generated subjects to remain consistent with the reference appearance. For camera control, one common practice~\cite{he2024cameractrl, he2025cameractrl2} is to convert the camera path into Plücker embeddings~\cite{sitzmann2021light} and inject them into the main network, guiding the synthesized video to follow the specified trajectory. In contrast, our model controls the camera by navigating through 3DGS views: given a 3DGS scene and a camera path, we render a projection video along the path, which conditions the generator to follow the desired motion faithfully.

\noindent\textbf{Memory Mechanism in Video Generation.} 
Recent works incorporate memory mechanisms to improve long-term spatial and temporal consistency in video generation. These approaches retrieve generation history to localize relevant content across modalities such as RGB~\cite{yu2025cam} and depth~\cite{chen2025deepverse}, often using surfel-indexed view selection~\cite{li2025vmem} or camera FOV overlap~\cite{xiao2025worldmem, yu2025cam}. Other methods~\cite{yu2024wonderjourney, huang2025voyager, wu2025videoworld, huang2025memoryforcing} maintain a global point cloud map during generation, enabling the model to identify and reuse the most relevant spatial regions, thereby maintaining coherence across continuously generated scenes. In our work, the video generation with memory mechanism is realized by conditioning on both the character and the 3DGS scene. The 3DGS scene serves as a spatial memory that explicitly encodes the geometric and appearance information of the environment, while the character provides dynamic cues for motion and interaction.

\noindent\textbf{World Models for Static Scene Creation.}
Existing world models that generate static yet explorable environments can be broadly categorized into two types. The first type~\cite{valevski2024gamengen, decart2024oasis, bruce2024genie, feng2024thematrix, yang2024playable, kanervisto2025world, zhang2025matrix1, yu2025gamefactory, sun2025realplay, li2025hunyuangamecraft, mao2025yume, he2025matrix, ye2025yan, rtfm, genie3, lingbot-world} stores the world implicitly within neural networks, using video generation models~\cite{yang2024cogvideox, kong2024hunyuanvideo, wan2025, veo} to visualize it. Users provide navigation commands (e.g., ``camera forward'') drawn from a predefined set of camera trajectories, and the model synthesizes new frames along this path while maintaining spatial consistency with past generations. The second type~\cite{shriram2024realmdreamer, li2024director3d} explicitly constructs a 3DGS world, where multi-view images are optimized to form a manipulable 3D representation, allowing users to render novel views from arbitrary camera poses. Further developments extend this paradigm by using panoramic inputs~\cite{zhang2025worldprompter, yang2025matrix3d} or directly generating 3DGS representations from text or a single image through feed-forward networks~\cite{yang2025prometheus, szymanowicz2025bolt3d, go2025splatflow, liang2025wonderland, liu2025worldmirror}, while others~\cite{sun2024dimensionx, chen2025flexworld, wang2025videoscene, zhang2025scenesplatter, hao2025gaussvideodreamer} integrate video generation to streamline and accelerate 3DGS creation. In this work, users can either create or specify a 3DGS scene before generation. When users do not provide one, we adopt Marble~\cite{worldlabs} to automatically generate a static 3DGS world.

\label{sec:related-works}

\section{Method}
\label{sec:method}

\begin{figure*}[!t]
    \centering
    \includegraphics[width=\linewidth]{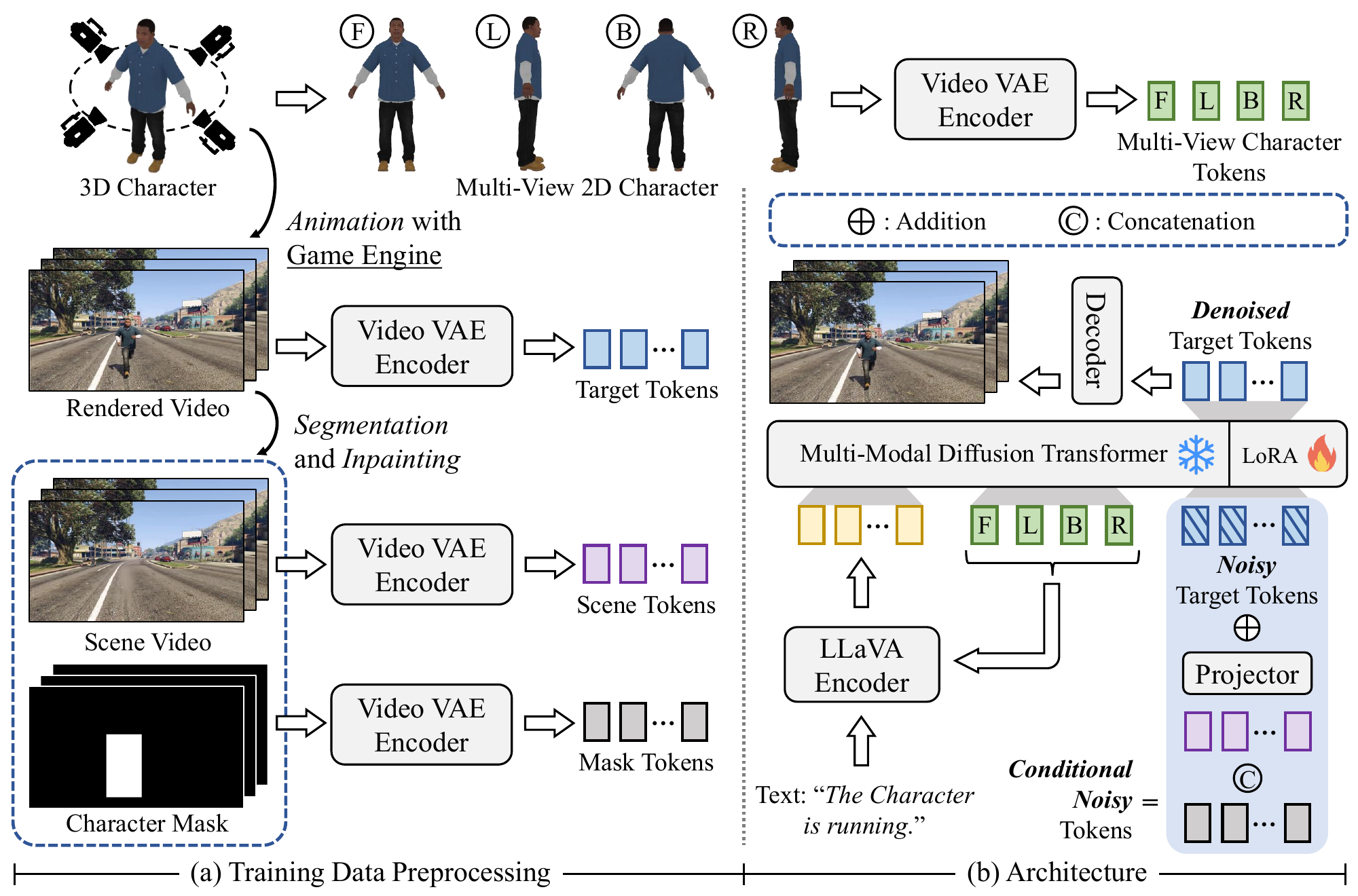}
    \caption{(a) Each training sample consists of a 3D character and a video depicting the character performing an action described by a short text. Through segmentation and inpainting, we obtain the corresponding scene video and character mask sequence. The VAE encoder is then applied to encode these inputs into tokens. (b) CustomX predicts target video tokens conditioned on scene, mask, text, and multi-view character tokens within a Multi-Modal Diffusion Transformer, trained using Flow Matching~\cite{lipman2022flow}. Refer to \cref{fig:AR} for the training process of the auto-regressive mode, which enables iterative interaction with CustomX, and \cref{fig:inference} for the inference.}
    \label{fig:overview}
\end{figure*}

In this section, we illustrate the detailed design of our method. Given a pre-generated or real-world reconstructed 3DGS scene $\mathcal{S}$ and a user-specified 3D character $\mathcal{C}$, \textbf{CustomX} enables users to iteratively control the character $\mathcal{C}$ via text instructions $\mathcal{T}$ within the scene $\mathcal{S}$, generating long-horizon, temporally coherent video clips that remain visually consistent with both $\mathcal{S}$ and $\mathcal{C}$.

We formulate this process as a conditional autoregressive video generation problem. At each iteration, CustomX synthesizes the current video clip conditioned on multiple multi-modal inputs: the previous generated clip, character representations, scene representations, and the current text instruction. Overviews of the training and inference pipelines are shown in \cref{fig:overview} and \cref{fig:inference}, respectively.

\subsection{Training Data Pre-Processing}
\label{sec:data-pre-processing}
\noindent\textbf{Training Set Construction.} As shown in \cref{fig:overview}(a), our training data is GTA-V~\cite{gtav}, a game where players can control a character to perform basic actions such as ``run forward''. We record gameplay sequences and segment them into short video clips, ensuring that each clip (1) contains only a single action and (2) has a fixed length of 129 frames. For each clip $\boldsymbol{V}$, we apply the following steps:
\begin{enumerate}
    \item \textit{Character Segmentation.} We use Grounded-SAM-2~\cite{ren2024groundedsam} to segment characters and extract their bounding-box mask sequences, denoted as $\boldsymbol{M}$.
    \item \textit{Scene Inpainting.} We remove the segmented characters and apply DiffuEraser \cite{li2025diffueraser} to fill the missing regions, yielding the inpainted scene video $\boldsymbol{S}$.
    \item \textit{Action Labeling.} Each clip is then annotated with a concise text label $\boldsymbol{T}$ describing the action performed by the character, such as ``The character is running forward''.
\end{enumerate}

GTA-V also provides access to 3D character models used in the game. To ease the character modeling, we represent each character using four viewpoint renders~\cite{zhang2024clay}—front, left, right, and back—denoted as $\boldsymbol{C} = \{\boldsymbol{C}_{F}, \allowbreak \boldsymbol{C}_{L}, \allowbreak \boldsymbol{C}_{R}, \allowbreak \boldsymbol{C}_{B}\}$. 

Finally, each processed training sample is represented as a tuple $\{\boldsymbol{V}, \allowbreak \boldsymbol{S}, \allowbreak \boldsymbol{M}, \allowbreak \boldsymbol{T}, \allowbreak \boldsymbol{C}\}$, forming a structured dataset.

\noindent\textbf{Token Extraction.} 
As shown in \cref{fig:overview}(a), given a training sample $\{\boldsymbol{V}, \allowbreak \boldsymbol{S}, \allowbreak \boldsymbol{M}, \allowbreak \boldsymbol{T}, \allowbreak \boldsymbol{C}\}$, we adopt the video VAE encoder from HunyuanCustom~\cite{hu2025hunyuancustom} to extract tokens for the video $\boldsymbol{V}$, scene $\boldsymbol{S}$, and mask $\boldsymbol{M}$. The resulting token sequences are denoted as $\mathcal{T}_{\boldsymbol{V}}$, $\mathcal{T}_{\boldsymbol{S}}$ and $\mathcal{T}_{\boldsymbol{M}}$, respectively. The video VAE encoder operates with a spatial downsampling rate of 8 and a temporal downsampling rate of 4.

Note that the video VAE encoder can also be applied to single images. Therefore, for the multi-view character $\boldsymbol{C} = \{\boldsymbol{C}_{F}, \boldsymbol{C}_{L},\boldsymbol{C}_{R},    \boldsymbol{C}_{B}\}$, we use the same encoder to extract tokens from each view, resulting in the multi-view character token set $\mathcal{T}_{\boldsymbol{C}} = \{\mathcal{T}_{\boldsymbol{C}_{F}}, \mathcal{T}_{\boldsymbol{C}_{L}},\mathcal{T}_{\boldsymbol{C}_{R}},\mathcal{T}_{\boldsymbol{C}_{B}}\}$.

Finally, to extract text tokens, following HunyuanCustom~\cite{hu2025hunyuancustom}, we employ the multi-modal encoder LLaVA~\cite{liu2023llava}, which takes both the text instruction $\boldsymbol{T}$ and character tokens $\mathcal{T}_{\boldsymbol{C}}$ as input. The resulting encoded text tokens are denoted as $\mathcal{T}_{\boldsymbol{T}}$. Implementation details are provided in the appendix. 

Now, the training sample $\{\boldsymbol{V}, \boldsymbol{S}, \boldsymbol{M}, \boldsymbol{T}, \boldsymbol{C}\}$ is fully encoded into the latent space as $\{\mathcal{T}_{\boldsymbol{V}}, \mathcal{T}_{\boldsymbol{S}}, \mathcal{T}_{\boldsymbol{M}}, \mathcal{T}_{\boldsymbol{T}}, \mathcal{T}_{\boldsymbol{C}}\}$.

\subsection{Architecture}
\label{sec:architecture}

\noindent\textbf{Training Objective.} \cref{fig:overview}(b) illustrates the architecture of CustomX, whose backbone consists of a stack of full-attention Transformer blocks. We adopt Flow Matching~\cite{lipman2022flow} for model training, conditioned on multiple inputs (i.e., $\mathcal{T}_{\boldsymbol{S}}$, $\mathcal{T}_{\boldsymbol{M}}$, $\mathcal{T}_{\boldsymbol{T}}$, and $\mathcal{T}_{\boldsymbol{C}}$), to guide the generation process from pure noise to $\mathcal{T}_{\boldsymbol{V}}$. 

Concretely, given target video tokens $\mathcal{T}_{\boldsymbol{V}}$, we first sample $t \in [0, 1]$ from a logit-normal distribution and initialize the noise $\mathbf{x}_0 \sim \mathcal{N}(\mathbf{0}, \mathbf{I})$ following Gaussian distribution. The intermediate sample $\mathbf{x}_t = (1-t) \mathbf{x}_0 + t \mathcal{T}_{\boldsymbol{V}}$ is then obtained via linear interpolation. The model is trained to predict the velocity $\mathbf{u}_t = d\mathbf{x}_t/dt$ by minimizing the mean squared error between the predicted velocity $\mathbf{v}_t$ and the ground-truth velocity $\mathbf{u}_t$:
\begin{equation}
\label{eq:loss}
    \mathcal{L} = \mathbb{E}_{t, \mathbf{x}_0, \mathcal{T}_{\boldsymbol{V}}} \left\| \mathbf{v}_t - \mathbf{u}_t \right\|^2.
\end{equation}

\noindent\textbf{Condition Modeling.} We incorporate multiple conditioning signals to guide the learning process, including text tokens $\mathcal{T}_{\boldsymbol{T}}$, multi-view character tokens $\mathcal{T}_{\boldsymbol{C}}$, scene tokens $\mathcal{T}_{\boldsymbol{S}}$, and mask tokens $\mathcal{T}_{\boldsymbol{M}}$. As illustrated in \cref{fig:overview}(b), to inject scene and mask priors, we directly fuse $\mathcal{T}_{\boldsymbol{S}}$ and $\mathcal{T}_{\boldsymbol{M}}$ into the noisy target tokens $\mathbf{x}_t$ via: $\mathbf{x}'_t = \mathbf{x}_t + \text{Projector}([\mathcal{T}_{\boldsymbol{S}}; \mathcal{T}_{\boldsymbol{M}}])$,
where $[~;~]$ denotes channel-wise concatenation. The $\text{Projector}$ maps the concatenated tokens $[\mathcal{T}_{\boldsymbol{S}}; \mathcal{T}_{\boldsymbol{M}}]$ to the same dimensionality as $\mathbf{x}_t$ using a lightweight linear layer. We refer to the resulting $\mathbf{x}'_t$ as the conditional noisy tokens.

At last, to integrate the text tokens $\mathcal{T}_{\boldsymbol{T}}$ and multi-view character tokens $\mathcal{T}_{\boldsymbol{C}}$, we concatenate $\mathcal{T}_{\boldsymbol{T}}$, $\mathcal{T}_{\boldsymbol{C}}$, and $\mathbf{x}'_t$ along the sequence dimension. The concatenated sequence is then fed into the backbone, which is implemented as a stack of full-attention Transformer blocks, to denoise $\mathbf{x}_t$ under the supervision of \cref{eq:loss}.

\begin{figure}[!t]
    \centering
    \includegraphics[width=0.5\linewidth]{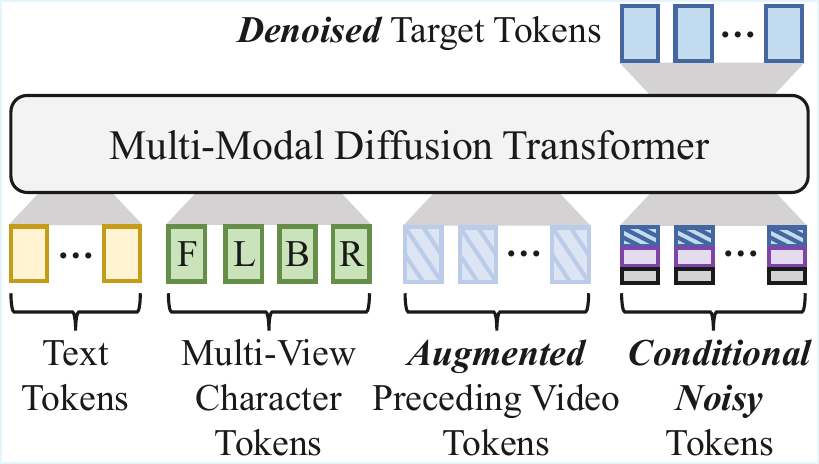}
    \caption{Illustration of the auto-regressive mode. The only difference from the original architecture in \cref{fig:overview} is the addition of an extra conditioning input, i.e., the preceding video tokens. Note that a misalignment exists between training and inference: during training, the preceding video tokens are derived from ground-truth videos, whereas during inference, they are generated by the model itself. To mitigate this discrepancy, we add a small Gaussian noise to the preceding video tokens during training and refer to the resulting tokens as augmented preceding video tokens.}
    \label{fig:AR}
\end{figure}

\noindent\textbf{Positional Embeddings.} Following HunyuanCustom~\cite{hu2025hunyuancustom}, no positional embeddings are added to the text tokens $\mathcal{T}_{\boldsymbol{T}}$. A standard 3D-RoPE (over time, height, and width dimensions) is applied to the conditional noisy tokens $\mathbf{x}'_t$. For the multi-view character tokens $\mathcal{T}_{\boldsymbol{C}} = \{\mathcal{T}_{\boldsymbol{C}_{F}}, \mathcal{T}_{\boldsymbol{C}_{L}},\mathcal{T}_{\boldsymbol{C}_{R}},\mathcal{T}_{\boldsymbol{C}_{B}}\}$, each $\mathcal{T}_{\boldsymbol{C}_{F}}$,  $\mathcal{T}_{\boldsymbol{C}_{L}}$, $\mathcal{T}_{\boldsymbol{C}_{R}}$, and $\mathcal{T}_{\boldsymbol{C}_{B}}$ represents single-view character tokens. For each view, a shifted-3D-RoPE is applied. Taking $\mathcal{T}_{\boldsymbol{C}_{F}}$ as an example,
\begin{equation} \text{PE}_{\mathcal{T}_{\boldsymbol{C}_{F}}}(i,j) = \text{3D-RoPE}(-1, i+w, j+h),
\end{equation}
where $(w,h)$ denotes the spatial size of $\mathcal{T}_{\boldsymbol{C}_{F}}$, and the shifts along the temporal and spatial dimensions are $-1$ and $(w,h)$, respectively. For $\mathcal{T}_{\boldsymbol{C}_{L}}$, $\mathcal{T}_{\boldsymbol{C}_{R}}$ and  $\mathcal{T}_{\boldsymbol{C}_{B}}$, the spatial shift remains the same, while the temporal shifts are set to $-2$, $-3$, and $-4$, respectively.

\begin{figure}[!t]
    \centering
    \includegraphics[width=\linewidth]{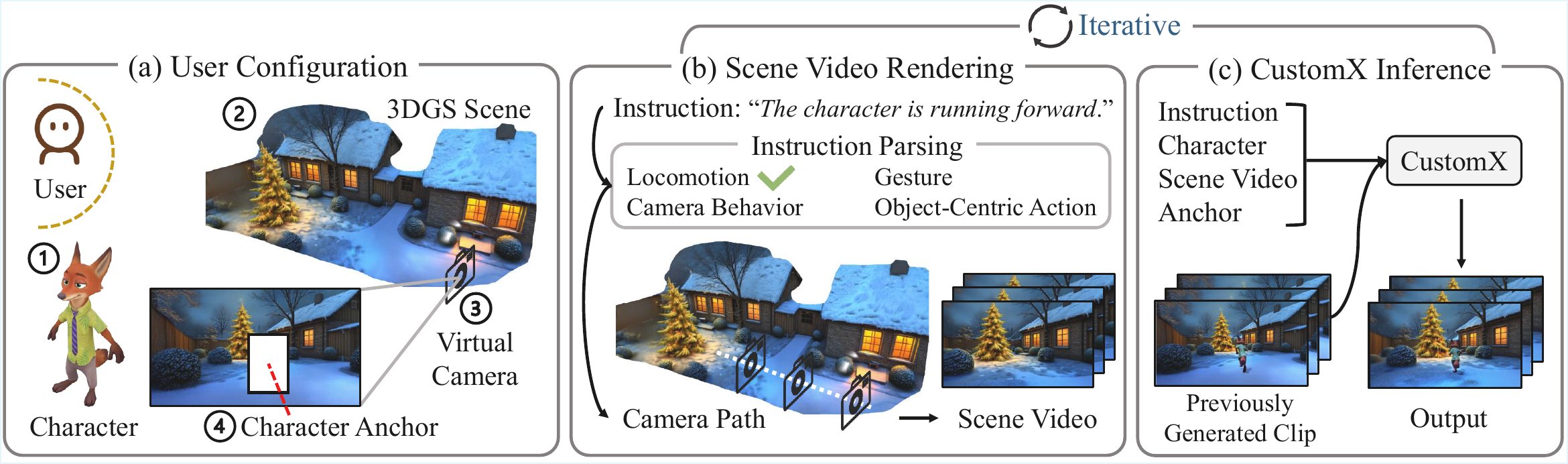}
    \caption{Inference of CustomX. (a) Users first specify the inputs, including the character, 3DGS scene, virtual camera location, and character anchor. (b) The user-provided text instruction is parsed, and a corresponding camera path is generated. Applying this path to the 3DGS scene produces a rendered scene video. (c) CustomX then takes multiple inputs as conditions to generate the final output. Steps (b) and (c) can be performed iteratively, enabling temporally consistent, long-horizon interactions.}
    \label{fig:inference}
\end{figure}

\subsection{Auto-Regressive Mode}
\label{sec:AR}
CustomX supports multi-round user interaction while maintaining temporal continuity and semantic coherence between adjacent video clips. To achieve this, we extend CustomX into an auto-regressive mode. Specifically, we divide the target video tokens $\mathcal{T}_{\boldsymbol{V}}$ along the temporal dimension into two parts: the first quarter, denoted as $\mathcal{T}_{\boldsymbol{V}_{1}}$, serves as the preceding video tokens, and the remaining three quarters, denoted as $\mathcal{T}_{\boldsymbol{V}_{[2:4]}}$, serve as the \textit{new} target video tokens. 

The model is trained to generate $\mathcal{T}_{\boldsymbol{V}_{[2:4]}}$ conditioned on both the preceding video tokens $\mathcal{T}_{\boldsymbol{V}_{1}}$ and the other conditioning signals introduced in \cref{sec:architecture}. As shown in \cref{fig:AR}, to incorporate the newly added condition $\mathcal{T}_{\boldsymbol{V}_{1}}$, we prepend it to the conditional noisy tokens. The fusion strategy for the other conditioning signals remains unchanged, as described in \cref{sec:architecture}.

\subsection{Inference and Acceleration}
\label{sec:inference}

\noindent\textbf{Inference.} \cref{fig:inference} illustrates the inference pipeline of CustomX, which consists of three main stages:
\begin{enumerate}
    \item \textit{User Configuration}. Users first specify a character and a 3DGS scene. They can place a virtual camera at any desired viewpoint within the scene and define a single-frame bounding-box mask (i.e., an anchor) to indicate where the character should appear in the generated video. This anchor is automatically propagated to all subsequent frames, forming per-frame anchors and keeping the manual effort minimal. Users may also leverage existing tools to create the 3D character (e.g., Hunyuan3D~\cite{lai2025hunyuan3d}) or the 3DGS scene (e.g., World Labs Marble~\cite{worldlabs}).

    \item \textit{Scene Video Rendering}. Next, users provide a text instruction. The instruction is parsed into four categories: (a) locomotion, (b) gesture, (c) object-centric action, and (d) camera behavior. Each category determines a different camera path. For (a), CustomX generates a camera trajectory consistent with the motion described in the text, for example, for ``The character is running forward'', the camera follows a forward-moving path. For (b) and (c), the camera remains stationary. For (d), CustomX generates a trajectory matching the specified camera motion, for instance, for ``The camera orbits around the character'', the camera follows a circular path around the character. CustomX then renders a scene video clip along the corresponding camera trajectory.
    \item \textit{CustomX Inference}. Finally, CustomX encodes the text instruction, character, scene, character anchor, and the previously generated clip (optional, for auto-regressive mode) into tokens using the encoders illustrated in \cref{fig:overview}. These tokens are then fed into the CustomX model to generate the final output. 
\end{enumerate}

Note that Steps 2 and 3 can be performed iteratively, enabling temporally consistent, long-horizon generation.

\noindent\textbf{Acceleration.} To accelerate inference, we adopt DMD2~\cite{yin2024improved} to distill the original 30-step denoising model into a more efficient 4-step version.
\section{Experiment}
\label{sec:experiment}

\subsection{Experimental Settings}
\label{sec:exp_settings}
\noindent\textbf{Training Details.} Our model is initialized from HunyuanCustom~\cite{hu2025hunyuancustom}, which contains 13B parameters. We freeze the LLaVA encoder and the scene condition projector, and apply LoRA-style~\cite{hu2022lora} fine-tuning to the backbone with a rank of 64. Two separate models are trained for 360P and 720P data using the AdamW optimizer with a learning rate of 1e-4, each for 5,000 steps under the ZeRO-2 strategy. The 360P model is trained on 8$\times$ NVIDIA H100 GPUs, while the 720P model, owing to its higher resolution, is trained on 8$\times$ NVIDIA B200 GPUs. Further details on training and acceleration are provided in the appendix.

\noindent\textbf{Training Data.}
Following the data curation procedure described in \cref{sec:data-pre-processing}, we construct a training set comprising 2,084 video samples featuring five characters. Each sample is annotated with text labels describing either locomotion actions (\{``run forward'', ``run to the left'', ``run to the right'', ``run backward''\}) or camera behaviors (\{``orbit'', ``follow'' \}). A key observation of CustomX is that post-training on such simple locomotion and camera-behavior data can substantially enhance pre-trained models, improving both action dynamics and camera control capability. This is analogous to large language models~\cite{guo2025deepseekr1, yang2025qwen3}, where extensive pre-training endows rich world knowledge, while post-training, which is conducted with a much smaller dataset, serves to activate specific capabilities (e.g., in GPT-3~\cite{ouyang2022instructgpt}, post-training data is orders of magnitude smaller than the pre-training corpus). For each sample, we prepare two resolutions, 360P and 720P, to train models of different quality levels. Unless otherwise specified, the 360P version is used by default. Additional details are provided in the appendix.

\noindent\textbf{Evaluation.}
We evaluate our model across four aspects: (1) visual quality, (2) camera controllability, (3) action control capability and generalization to novel actions, and (4) character consistency on novel characters. For (1) and (2), we adopt the WorldScore~\cite{duan2025worldscore} metrics to assess the generated samples. For (3), we measure the control success rate via human evaluation and report the CLIP~\cite{radford2021clip} text-to-image similarity score, covering both the four seen locomotion actions and up to 142 novel actions. For (4), we assess character similarity between the ground-truth character and the generated one using 30 novel characters, evaluated by DINOv2~\cite{oquab2023dinov2} and CLIP~\cite{radford2021clip} scores. By default, we use the 360P version of our model. At each iteration, our model generates 96 frames, using the previously generated 33 frames as conditions when available. Unless otherwise noted, evaluations are conducted on the first generated clip.

\begin{table*}[!t]
\caption{WorldScore metrics for evaluating generation quality, categorized into three groups: (1) controllability, (2) quality, and (3) dynamics. The static and dynamic scores are computed by aggregating metrics from these three groups. $\dagger$ denotes dedicated world models. \textit{Ctrl}: Controllability; \textit{Align}: Alignment; \textit{Const}: Consistency; \textit{Photo}: Photometric; \textit{Acc}: Accuracy; \textit{Mag}: Magnitude; \textit{Smooth}: Smoothness.}
\resizebox{\linewidth}{!}{%
\centering
\small
\setlength{\tabcolsep}{3pt}
\begin{tabular}{@{}l|*{11}{c|}c@{}}
\toprule
\multirow{4}{*}{Method} & \multicolumn{2}{c|}{WorldScore} &
\multicolumn{3}{c|}{Controllability} &
\multicolumn{4}{c|}{Quality} &
\multicolumn{3}{c}{Dynamics} \\ \cmidrule(lr){2-3}
\cmidrule(lr){4-6} \cmidrule(lr){7-10} \cmidrule(lr){11-13}
& \makecell{Static} & \makecell{Dyna-\\mic} &
\makecell{Camera\\Ctrl} &
\makecell{Object\\Ctrl} &
\makecell{Content\\Align} &
\makecell{3D\\Const} &
\makecell{Photo\\Const} &
\makecell{Style\\Const} &
\makecell{Subjective\\Quality} &
\makecell{Motion\\Acc} &
\makecell{Motion\\Mag} &
\makecell{Motion\\Smooth} \\
\midrule
CogVideoX1.5-I2V (5B)~\cite{yang2024cogvideox} & {60.08} & {56.77} & {42.13} & \textbf{100.00} & {31.12} & {68.65} & {81.43} & {77.86} & {19.39} & {54.92} & {24.58} & {67.62} \\
HunyuanVideo-I2V (13B)~\cite{kong2024hunyuanvideo} & {56.43} & {55.14} & {27.30} & \textbf{100.00} & {13.70} & {58.24} & \textbf{93.22} & {91.97} & {10.58} & {59.36} & {24.46} & {72.58} \\
Wan2.1-I2V (14B)~\cite{wan2025} & {57.91} & {55.87} & {37.32} & {98.33} & {26.34} & {81.88} & {83.77} & {65.03} & {12.73} & {59.59} & {28.60} & {65.07} \\
Wan2.2-I2V (14B)~\cite{wan2025} & {54.52} & {51.74} & {24.79} & \textbf{100.00} & {24.03} & {57.95} & {59.44} & \underline{98.91} & {16.51} & \underline{59.60} & \underline{38.93} & {37.26} \\
Wan-VACE (14B)~\cite{vace} & {51.54} & {52.03} & {21.29} & \textbf{100.00} & {30.39} & {27.53} & {54.18} & \textbf{99.02} & {17.39} & {56.73} & {34.78} & {67.98} \\
Wan-VACE (1.3B)~\cite{vace} & {50.75} & {49.38} & {31.12} & \textbf{100.00} & {11.54} & {29.66} & {61.00} & {97.67} & {24.26} & {55.87} & {29.48} & {53.17} \\
HunyuanCustom (13B)~\cite{hu2025hunyuancustom} & {62.64} & \underline{61.11} & {47.19} & \textbf{100.00} & \underline{72.07} & {48.06} & {31.40} & {97.47} & \underline{25.84} & {59.04} & {24.05} & \underline{89.56} \\
DeepVerse$^\dagger$~\cite{chen2025deepverse} & {52.63} & {47.63} & {52.48} & {75.00} & {18.80} & {35.47} & {83.16} & {92.39} & {11.09} & {33.30} & {33.61} & {40.97} \\
Hunyuan-GameCraft$^\dagger$~\cite{li2025hunyuangamecraft} & \underline{69.92} & {57.77} & \underline{77.45} & {83.33} & {51.16} & \textbf{85.91} & {82.12} & {93.39} & {16.11} & {16.65} & {31.83} & {39.77} \\
Matrix-Game-2.0$^\dagger$~\cite{he2025matrix} & {52.26} & {43.98} & {15.10} & \underline{99.17} & {12.38} & {60.29} & {68.35} & {97.60} & {12.92} & {3.07} & \textbf{47.36} & {23.59} \\
\midrule
CustomX (Ours) & \textbf{84.64} & \textbf{77.22} & \textbf{88.91} & \textbf{100.00} & \textbf{75.73} & \underline{83.57} & \underline{87.68} & \underline{98.91} & \textbf{57.72} & \textbf{61.08} & {24.12} & \textbf{94.47} \\
\bottomrule
\end{tabular}%
}
\label{tab:worldscore}
\end{table*}

\subsection{Main Results}
\label{sec:main}
\noindent\textbf{Visual Quality Evaluation.}
We adopt the metrics proposed by WorldScore~\cite{duan2025worldscore} to evaluate the visual quality of generated videos. WorldScore defines a suite of metrics categorized into three groups: \textit{Controllability}, \textit{Quality}, and \textit{Dynamics}. The first two primarily assess visual fidelity in static regions, while the last evaluates motion quality in dynamic regions (i.e., the character in our case).

We compare our model with dedicated world models, including DeepVerse~\cite{chen2025deepverse}, Hunyuan-GameCraft~\cite{li2025hunyuangamecraft}, and Matrix-Game-2.0~\cite{he2025matrix}, that focus on controlling the main entity, as well as with video generation base models listed in \cref{tab:worldscore}.

\begin{table*}[!t]
\caption{Action control and character consistency evaluation on general foundation models and dedicated world models$^\dagger$. Locomotion actions include ``run forward'', ``run to the left'', ``run to the right'', and ``run backward'', while richer actions encompass 142 gesture and object-centric actions. Note that the three world models restrict action control to locomotion only. }
\resizebox{\linewidth}{!}{%
\centering
\small
    \centering
    \begin{tabular}{l|c|c|c|c|c|c}
        \toprule
        \multirow{5}{*}{Method} & \multicolumn{4}{c|}{Action Control} & \multicolumn{2}{c}{Character Consistency} \\ 
        \cmidrule(lr){2-5} \cmidrule(lr){6-7} 
        
        & \multicolumn{2}{c|}{Success Rate (\%)} & \multicolumn{2}{c|}{CLIP Text-Image Score} 
        & \multirow{4}{*}{\makecell{DINOv2\\Score}} 
        & \multirow{4}{*}{\makecell{CLIP\\Score}} \\ 
        \cmidrule(lr){2-3} \cmidrule(lr){4-5}
        
        & \makecell{Locomotion\\Actions} & \makecell{Richer\\Actions}
        & \makecell{Locomotion\\Actions} & \makecell{Richer\\Actions}
        & & \\ 
        
        \midrule
        CogVideoX1.5-I2V (5B)~\cite{yang2024cogvideox} & 23.3 & 21.1 & 0.261 & 0.273 & 0.594 & 0.611 \\
        HunyuanVideo-I2V (13B)~\cite{kong2024hunyuanvideo} & 26.7 & 35.2 & 0.272 & 0.293 & 0.645 & \underline{0.709} \\
        Wan2.1-I2V (14B)~\cite{wan2025} & 26.7 & 64.8 & 0.267 & 0.302 & 0.627 & 0.678 \\
        Wan2.2-I2V (14B)~\cite{wan2025} & 53.3 & \underline{74.6} & 0.272 & \underline{0.303} & \underline{0.650} & 0.704 \\
        Wan-VACE (14B)~\cite{vace} & 43.3 & 73.2 & 0.270 & \underline{0.303} & 0.398 & 0.541 \\
        Wan-VACE (1.3B)~\cite{vace} & 26.7 & 13.4 & 0.261 & 0.269 & 0.504 & 0.548 \\
        HunyuanCustom (13B)~\cite{hu2025hunyuancustom} & \underline{56.7} & 51.4 & \underline{0.273} & 0.297 & 0.558 & 0.665 \\
        DeepVerse$^\dagger$~\cite{chen2025deepverse} & 6.7 & - & 0.259 & - & 0.291 & 0.523 \\
        Hunyuan-GameCraft$^\dagger$~\cite{li2025hunyuangamecraft} & 16.7 & - & 0.255 & - & 0.329 & 0.529 \\
        Matrix-Game-2.0$^\dagger$~\cite{he2025matrix} & 3.3 & - & 0.255 & - & 0.339 & 0.524 \\
        \midrule
        CustomX (Ours) & \textbf{100.0} & \textbf{80.7} & \textbf{0.279} & \textbf{0.305} & \textbf{0.698} & \textbf{0.721} \\
        \bottomrule
    \end{tabular}
    }
    \label{tab:compare_sota}
\end{table*}

For each model, we use the first two metric groups to evaluate 60 generated videos covering 30 different characters and two camera behaviors, \{``orbit'',``follow''\}. The third metric group is evaluated on 146 videos, where each video features the same character performing a distinct action, either one of four locomotion actions \{``run forward'', ``run to the right'', ``run to the left'', ``run backward''\} or one of 142 novel actions unseen during training. Note that (1) to the best of our knowledge, our model is the first framework to jointly integrate 3D or multi-view characters, action text prompts, scenes, and long-horizon generation, which necessitates an input configuration that naturally diverges from established baselines; (2) both foundation and dedicated models are evaluated using their original control mechanisms for character and camera, without any modification; (3) the three dedicated world models only support locomotion actions, thus they are evaluated solely on those; and (4) for models requiring an initial image input, we use Google Gemini~\cite{gemini} to render the corresponding character within the scene as the initialization. The results are shown in \cref{tab:worldscore}.

\noindent\textbf{Action Control and Generalization.} In \cref{tab:compare_sota}, we evaluate the model’s action control capability on both seen actions (four locomotion actions) and 142 novel actions (referred to as ``richer actions''). The novel actions cover both gesture-based behaviors (e.g., \textit{``wave hands''}) and object-centric interactions (e.g., \textit{``play a guitar''}). For each model, we conduct 30 evaluations on locomotion actions and 142 evaluations on distinct novel actions using the same set of characters. We report the action control success rate via human evaluation and the CLIP text-image similarity score, computed as the average frame-wise text-image score.

The results in \cref{tab:compare_sota} reveal that our model outperforms the base model, HunyuanCustom~\cite{hu2025hunyuancustom}. This phenomenon can be interpreted through the lens of post-training in large language models~\cite{ouyang2022instructgpt, rafailov2023dpo, guo2025deepseekr1, yang2025qwen3}, where fine-tuning typically does not disrupt the pre-trained representation space; rather, it adjusts the \textit{response style}, for example, to make the outputs more helpful or harmless, while preserving the extensive knowledge acquired during pre-training. In our case, the structurally simple fine-tuning data, composed primarily of fundamental locomotion behaviors, serve to refine \textit{motion dynamics} and align \textit{human embodiment representations}, rather than to redefine the model’s generative space. Consequently, our fine-tuning strategy enhances motion fidelity and behavioral coherence while maintaining the broad semantic and generative generalization inherited from large-scale pre-training.

\noindent\textbf{Character Consistency Evaluation.} A key feature of our model is its ability to maintain consistent visual identity between the provided character and the one appearing in the generated videos. In \cref{tab:compare_sota}, we evaluate character consistency using DINOv2 and CLIP scores. Both metrics measure the similarity between the generated and ground-truth characters. During evaluation, we crop the character region from each generated frame and compute the similarity score; the final result is averaged across all frames. Since our method takes multi-view inputs, each frame is scored by its maximum similarity to the ground-truth views. Each model is evaluated 30 times with different characters performing locomotion.

\begin{figure}[!t]
    \centering
    \includegraphics[width=\linewidth]{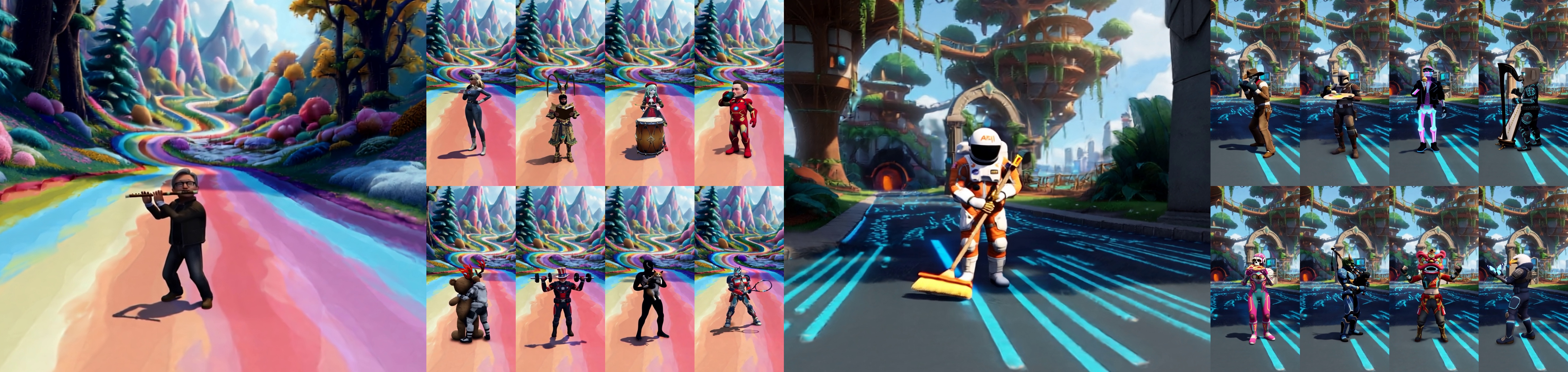}
    \caption{Screenshot visualizations of videos generated by CustomX, showcasing different characters performing various novel actions across two scenes. Additional examples are provided in the appendix and supplementary videos.}
    \label{fig:visualization}
\end{figure}

\begin{figure}[!t]
    \centering
    \includegraphics[width=\linewidth]{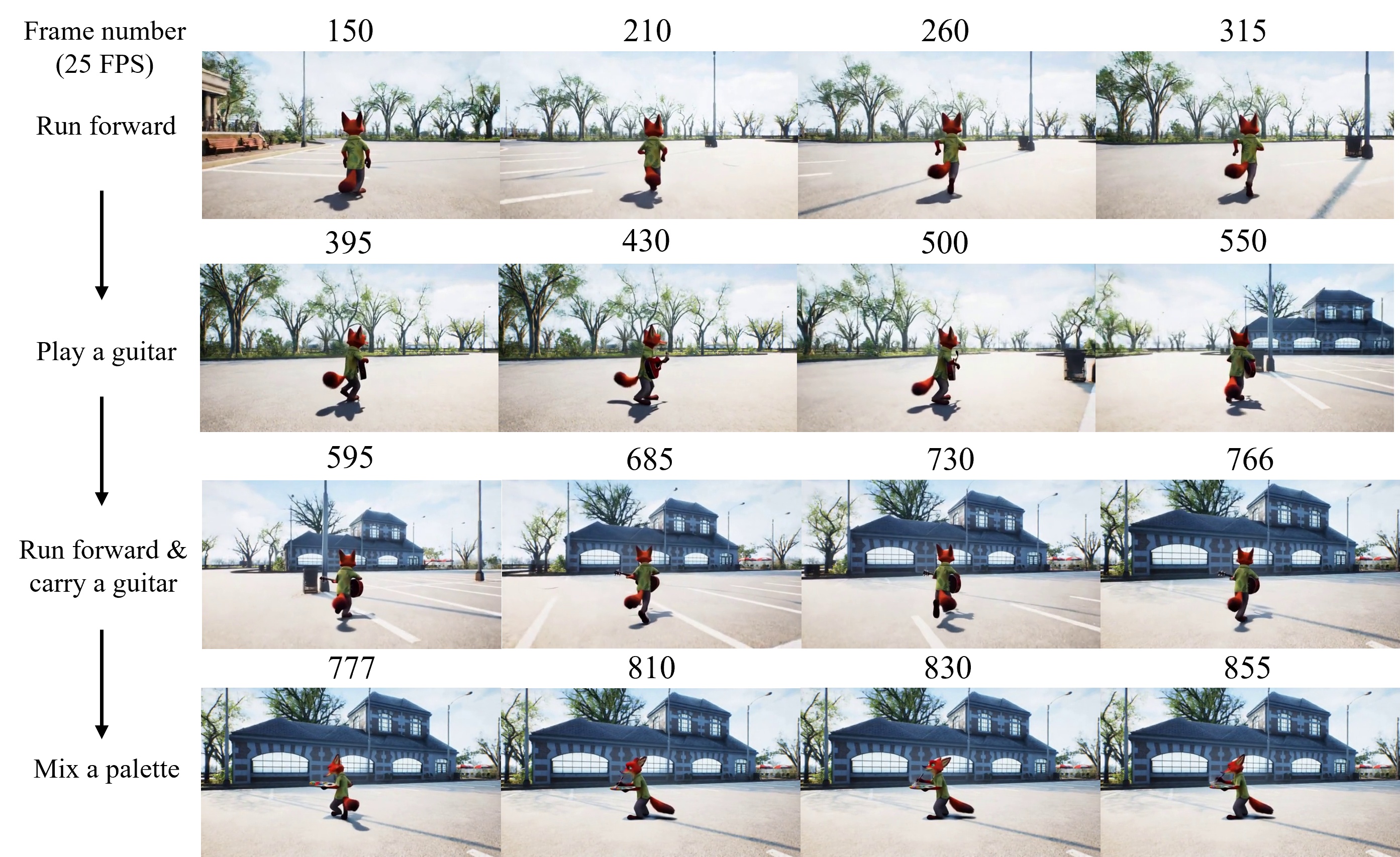}
    \caption{Screenshot visualizations of a long video generated by CustomX, showcasing a character iteratively exploring a scene with diverse actions. Additional long-form results are available in the appendix and supplementary videos.}
    \label{fig:visualization_long}
\end{figure}

\noindent\textbf{Visualization.} \cref{fig:visualization} shows screenshots from the generated videos, featuring different characters performing various actions across diverse scenes. \cref{fig:visualization_long} illustrates a long-horizon generation example in which users interact with the model four times. The resulting video follows the specified character, scene, and text instructions, while maintaining temporal coherence across generated clips. Additional examples are provided in the appendix and supplementary videos.


\subsection{Ablation Study}
\label{sec:ablation}

\noindent\textbf{Multi-View Character Condition.}
We compare our four-view character-condition model with two baselines: a single-view model and a front–back-view model. To evaluate character consistency, we generate videos by instructing characters to run toward the front, left, right, and back, which naturally reveals their appearance from multiple viewpoints. 
We then compute DINOv2 and CLIP scores following the protocol in \cref{sec:main}. We use the same dataset as that used in \cref{tab:compare_sota}. As shown in \cref{tab:multi_view_ablation}, character consistency improves as more view inputs are used.


\begin{table}[!t]
    \centering
    \small
    \begin{minipage}{0.42\textwidth}
        \centering
        \caption{Using multi-view character inputs improves character consistency. By default, we employ all four views.}
        \resizebox{\linewidth}{!}{
            \begin{tabular}{l|c|c}
                \toprule
                Character View & DINOv2 Score & CLIP Score \\
                \midrule
                Front & 0.556 & 0.628 \\
                ~+Back & 0.613 & 0.678 \\
                \midrule
                ~~+Right and Left & \textbf{0.698} & \textbf{0.721} \\
                \bottomrule
            \end{tabular}
        }
        \label{tab:multi_view_ablation}
    \end{minipage}
    \hfill
    \begin{minipage}{0.52\textwidth}
        \centering
        \caption{Using per-frame character anchors helps the model distinguish dynamic entities from the static scene, leading to higher DINOv2 and CLIP character consistency scores.}
        \resizebox{\linewidth}{!}{
            \begin{tabular}{l|c|c}
                \toprule
                Character Anchor Type & DINOv2 Score & CLIP Score \\
                \midrule
                w/o Anchor & 0.477 & 0.529 \\
                w/ First-Frame Anchor & 0.597 & 0.645 \\
                \midrule
                w/ Per-Frame Anchor & \textbf{0.698} & \textbf{0.721} \\
                \bottomrule
            \end{tabular}
        }
        \label{tab:mask_ablation}
    \end{minipage}
\end{table}

\noindent\textbf{Character Anchor Condition.} As illustrated in \cref{fig:overview}, we introduce an additional condition, namely a character mask, to enable the model to distinguish dynamic entities from static scenes. In the default training setting, a bounding-box mask, referred to as an anchor, is extracted for each frame. During inference, as shown in \cref{fig:inference}, users only need to specify a single-frame anchor, which is then propagated to all subsequent frames to form per-frame anchors. \cref{tab:mask_ablation} compares our default model, which uses per-frame anchors, with two variants: (1) a model without any anchor during both training and inference, and (2) a model that uses only the first-frame anchor, without propagating it to subsequent frames, throughout training and inference.

\begin{figure}[!t]
    \centering
    \small 
    \begin{minipage}[t]{0.37\textwidth}
        \centering
        \includegraphics[width=\linewidth]{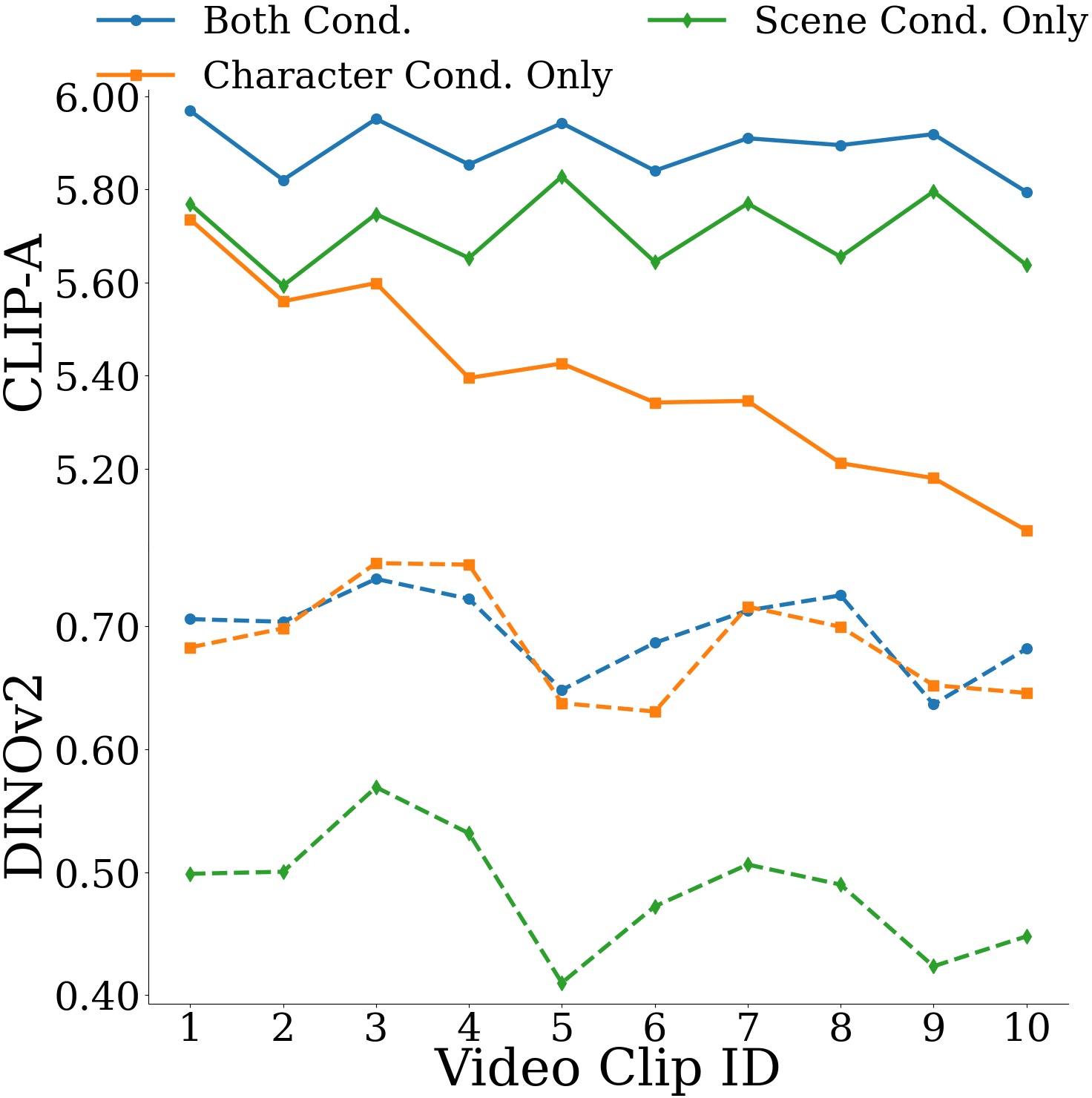}
        \caption{Using both visual conditions, namely the 3DGS scene and the multi-view character, significantly improves long-horizon visual quality across diverse video clips.}
        \label{fig:clip_aes}
    \end{minipage}
    \hfill
    \begin{minipage}[t]{0.6\textwidth}
        \centering
        \includegraphics[width=\linewidth]{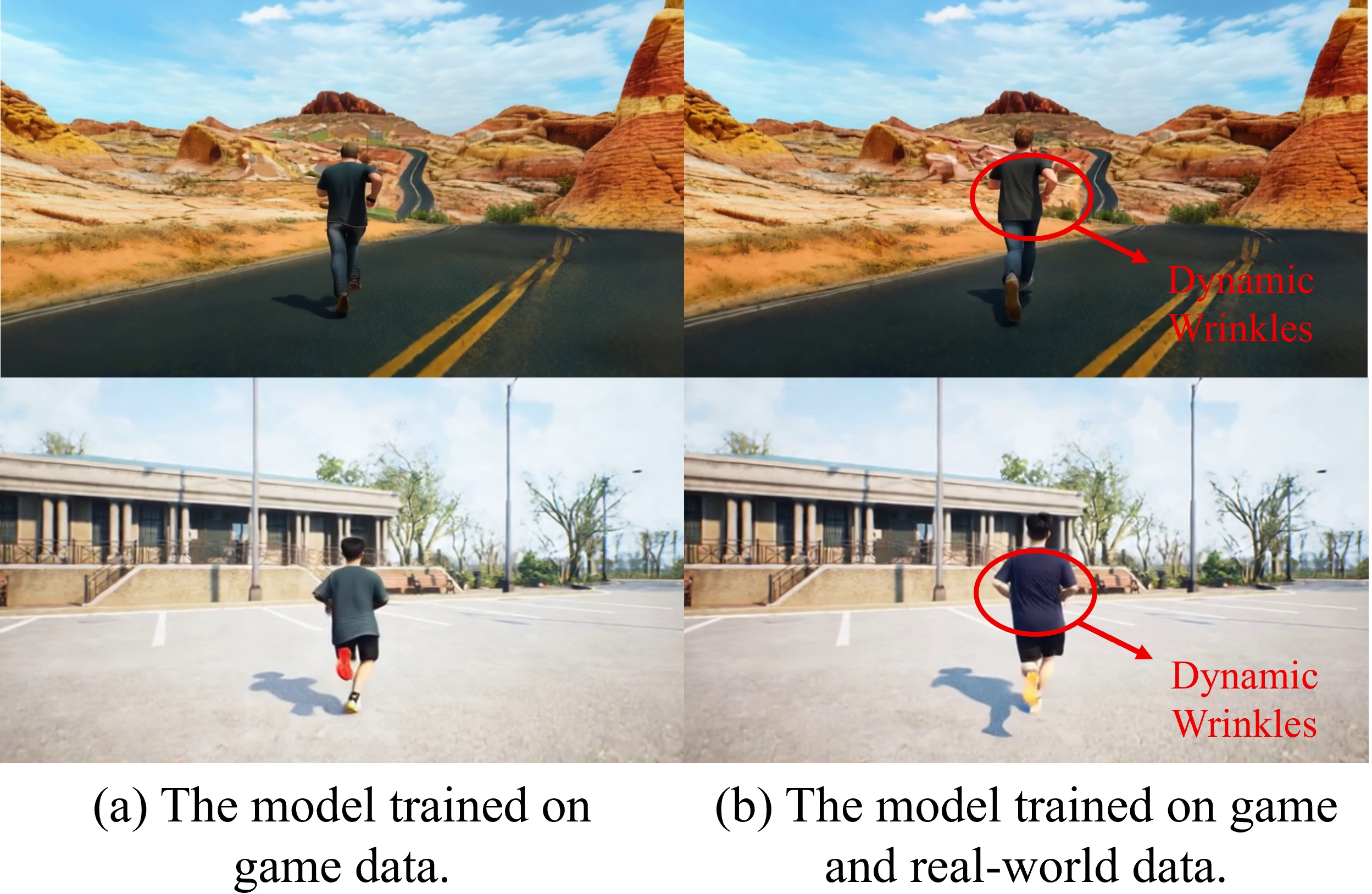}
        \caption{Evaluation of the hybrid data training strategy. (a) Training solely on game data causes the model to inherit a game-engine rendering style in synthesized characters. (b) Incorporating real-world data improves photorealism and enables the model to capture physical effects, such as dynamic clothing wrinkles, which are absent in GTA-V.}
        \label{fig:game_real_ablation_main}
    \end{minipage}
\end{figure}

\noindent\textbf{Visual Conditions Enhance Long-Horizon Generation.} We introduce two types of visual conditions, multi-view character and 3DGS scene, to ensure both character and scene consistency during generation. Beyond improving spatial coherence, these visual conditions also alleviate the issue of visual quality degradation over long-horizon generation. To validate this, we compare CustomX operating in the auto-regressive mode (see \cref{sec:AR}) against two variants: (1) using only the multi-view character condition, where the 3DGS scene condition is replaced with textual scene descriptions; and (2) using only the 3DGS scene condition, where the multi-view character condition is replaced with textual character descriptions. \cref{fig:clip_aes} presents a comparison with the two variants, using the CLIP-Aesthetic and DINOv2 scores as metrics to evaluate visual quality and character consistency, respectively. The evaluation protocol consists of 10 trials per model. In each trial, a different character is used to autoregressively generate 10 video clips, forming an iterative long-horizon generation process.

\noindent\textbf{Game-Real Hybrid Data Enhances Real-World Character Fidelity and Physics.} CustomX trained solely on GTA-V videos tends to inherit the game-engine rendering style, leading to stylized outputs even when conditioned on real-world characters. We therefore investigate whether incorporating real-world data into training can improve generation realism. Specifically, we collect additional 400 real-world videos using the same data collection pipeline as for GTA-V, except that the videos are captured from the real world rather than rendered by a game engine. As shown in \cref{fig:game_real_ablation_main}, the model trained on the hybrid dataset achieves higher realism and captures physical effects present in the real world but absent in games, such as dynamic clothing wrinkles. Additional details of the real-world dataset and further results are provided in the appendix.

\noindent\textbf{Acceleration.} Our 13B-parameter base model generates a 93-frame 360P video in 121s on a single NVIDIA H100 using a 30-step denoising schedule. By applying DMD2~\cite{yin2024improved}, we distill it into a 4-step version, reducing latency to 21s with only slight drops in DINOv2 (0.698$\rightarrow$0.669) and CLIP-Aesthetic (5.665$\rightarrow$5.583) scores. Latencies for higher resolutions are provided in the appendix.
\section{Conclusion}
\label{sec:conclusion}

We present \textit{CustomX}, a novel framework that allows users to provide a character and a 3DGS scene, enabling iterative interaction for both character control and world exploration. Unlike prior controllable-entity models that restrict the agent to a small set of predefined actions, CustomX supports open-ended control over a wide range of behaviors through natural language commands. CustomX delivers substantial improvements in motion dynamics and character consistency over its base model, as validated across a broad set of metrics including visual quality, action controllability, character fidelity, and long-horizon generation capability.

\noindent\textbf{Limitations.} As a controllable video generation framework, CustomX shares common limitations with existing video generation models. Although inference has been accelerated, it still does not operate in real time. Furthermore, the current model is limited to single-subject scenarios and cannot generate videos involving multiple simultaneously active characters.

\section*{Acknowledgements}
This work is supported in part by Microsoft Research, the HKU Startup Fund, and HKU Musketeers Foundation Institute of Data Science. 

\clearpage 
\begingroup \centering {\Large\bfseries Supplementary Material\par} \endgroup
\setcounter{section}{0} \setcounter{figure}{0} \setcounter{table}{0} \setcounter{equation}{0} 
\renewcommand{\thesection}{\Alph{section}} 
\renewcommand{\thefigure}{A\arabic{figure}} \renewcommand{\thetable}{A\arabic{table}} \renewcommand{\theequation}{A\arabic{equation}}

\section{More Implementation Details}

\subsection{Text Token Extraction}
Following HunyuanCustom~\cite{hu2025hunyuancustom}, we use the multi-modal encoder LLaVA~\cite{liu2023llava} to extract text tokens while incorporating the multi-view character images. Concretely, given the text instruction $\boldsymbol{T}$ and the character views ${\boldsymbol{C}}=\{{\boldsymbol{C}_{F}}, \allowbreak {\boldsymbol{C}_{L}}, \allowbreak {\boldsymbol{C}_{R}}, \allowbreak {\boldsymbol{C}_{B}}\}$, we construct the following prompt:
\begin{quote}
\small
“A character is [\textit{Action}]. \textless SEP\textgreater
The character front view looks like $\mathcal{E}({\boldsymbol{C}_{F}})$. \textless SEP\textgreater
The character left view looks like $\mathcal{E}({\boldsymbol{C}_{L}})$. \textless SEP\textgreater
The character back view looks like $\mathcal{E}({\boldsymbol{C}_{B}})$. \textless SEP\textgreater
The character right view looks like $\mathcal{E}({\boldsymbol{C}_{R}})$.”
\end{quote}
Here, [\textit{Action}] denotes the action description, $\mathcal{E}(\cdot)$ is the LLaVA image encoder, and \textless SEP\textgreater is the separation token used to distinguish text and visual modalities.

\subsection{More Training Details}
The model is initialized from the pre-trained weights of HunyuanCustom~\cite{hu2025hunyuancustom}, a multi-modal subject-driven variant of HunyuanVideo~\cite{kong2024hunyuanvideo}. Core components, which include the LLaVA encoder, the scene-condition projector, and the MMDiT \cite{peebles2023dit, esser2024mmdit} backbone, are kept frozen. Trainable parameters are introduced by injecting LoRA modules~\cite{hu2022lora} with a rank of 64 into the attention query, key, value, and projection matrices, as well as into the fully connected layers. We optimize the model using AdamW~\cite{loshchilov2017adamw} with a learning rate of 1e-4 and 500 warm-up steps. The scene condition is randomly dropped with a probability of 0.3, encouraging the model to rely more heavily on text and multi-view character references.

Our model supports two generation modes:
(1) First-clip generation. The model generates 93 frames, corresponding to 24 video latents (because the VAE encoder temporally downsamples a video of $N$ frames into $(N-1)/4+1$ video latents). In this setting, no preceding clip is provided. (2) Auto-regressive clip generation. When previous clips already exist, the model conditions on the last 33 frames (9 video latents) of the preceding clip and generates 96 new frames (24 video latents).

\subsection{Inference Acceleration}
To accelerate inference, we adopt the DMD2 distillation framework~\cite{yin2024improved}, employing teacher, student, and fake-score models initialized from our trained model. The teacher model remains fully frozen, while the student and fake-score models are fine-tuned using LoRA modules with a rank of 64. Following the DMD2 protocol, the fake-score model is updated every iteration, whereas the student model is updated once every five iterations. This setup—instantiating three 13B models while training two sets of LoRA parameters—incurs a substantial GPU memory overhead. As a result, DMD2 distillation is applied only to the 360P model and is trained for 4,000 steps on 8$\times$ NVIDIA B200 GPUs, using the ZeRO-3 optimization strategy to manage memory efficiently.

\section{More Experiments}

\begin{table}[!t]
\caption{Hybrid training with game and real-world data helps the model disentangle game-engine rendering from real-world visual characteristics, yielding higher DINOv2 and CLIP character consistency scores.}
    \centering
    \begin{tabular}{l|c|c}
        \toprule
        Training Data Type & DINOv2 Score & CLIP Score \\
        \midrule
        Game Data & 0.686 & 0.718 \\
        \midrule
        ~+Real-World Data & \textbf{0.755} & \textbf{0.729} \\
        \bottomrule
    \end{tabular}
    \label{tab:game_real_ablation}
\end{table}

\subsection{Game-Real Hybrid Data Enhances Real-World Character Fidelity and Physics}
As described in the main paper, training solely on GTA-V~\cite{gtav} videos introduces a challenge: the model tends to inherit the game-engine rendering style, causing synthesized characters to appear stylized even when conditioned on real-world multi-view characters at inference time. To improve the realism of generated real-world characters—while retaining the diverse and dynamic motion patterns learned from game data—we curate an additional real-world dataset and jointly train CustomX on both sources. Specifically, we record 400 video clips of real individuals performing the same set of locomotion actions as in the game dataset. These videos are processed using the same pipeline described in the main paper and standardized to 360P resolution.

The model is then trained on a hybrid dataset combining the aforementioned GTA-V and newly collected real-world videos. To help the model differentiate between rendered and real-world visual styles, we apply a simple data-tagging strategy~\cite{zhao2025synthetic}: GTA-V samples are tagged with the keyword ``rendered'' (e.g., ``The rendered character is running forward''), while real-world samples are tagged with ``real'' (e.g., ``The real character is running forward''). Aside from this tagging mechanism, the overall training procedure remains identical to that described in the main paper.

To evaluate the effectiveness of the mixed-data strategy, we collect multi-view captures of two unseen real-world individuals. Following the evaluation protocol in the main paper, we compute DINOv2~\cite{oquab2023dinov2} and CLIP~\cite{radford2021clip} scores to evaluate the character consistency. As shown in \cref{tab:game_real_ablation}, training on hybrid data produces measurable improvements in real-world character fidelity.

\subsection{Inference Acceleration}
\begin{figure}[!t]
    \centering
    \includegraphics[width=\linewidth]{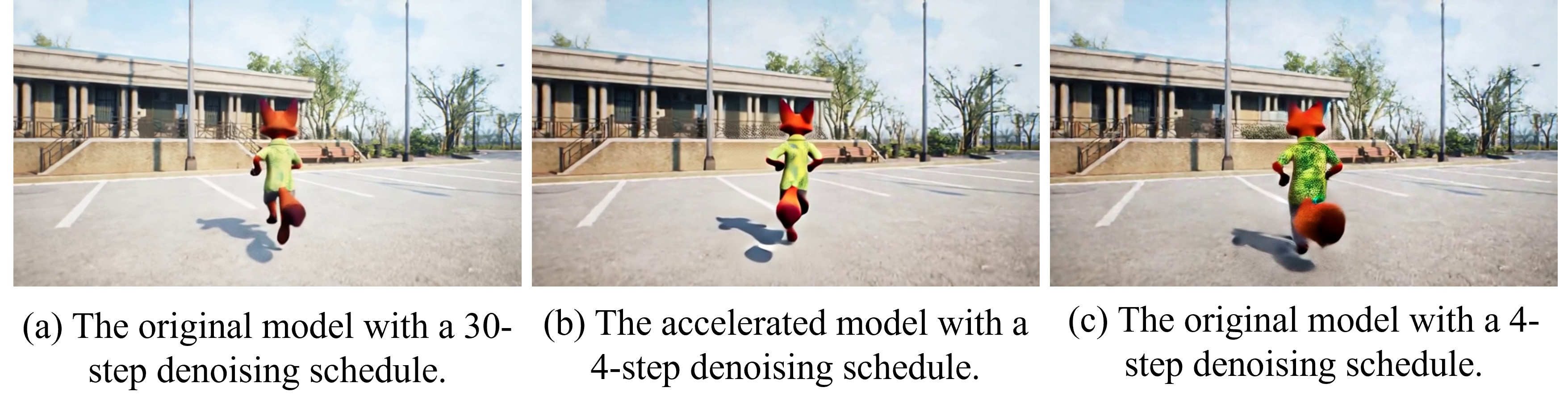}
    \caption{Qualitative comparison of three models: (a) the original model with a 30-step denoising schedule (no acceleration), (b) the accelerated model with a 4-step schedule, and (c) the original model restricted to 4 steps (no acceleration but fewer steps). The results show that our 4-step model matches the visual quality of the original model while achieving a $7.5\times$ inference speedup.}
    \label{fig:distill_comparison}
\end{figure}

To qualitatively assess the effectiveness of our inference acceleration strategy, \cref{fig:distill_comparison} compares three variants: (1) the original model with a 30-step denoising schedule (no acceleration), (2) the accelerated model with a 4-step denoising schedule, and (3) the original model also restricted to 4 denoising steps (no acceleration but fewer steps).
As illustrated in \cref{fig:distill_comparison}(a) and \cref{fig:distill_comparison}(b), our 4-step distilled model maintains visual fidelity on par with the original framework, while facilitating a $7.5\times$ inference speedup. Conversely, executing the non-accelerated original model with a simple 4-step denoising results in a pronounced degradation of visual quality, as evidenced in \cref{fig:distill_comparison}(c).


\subsection{Latencies for Higher-Resolution Inference}
\label{sec:inf_acc_supp}
We further report the inference cost and performance for generating higher-resolution outputs using 8$\times$ NVIDIA H100 GPUs. When producing a 93-frame video clip at 720P, the 720P model outperforms the 360P model in both DINOv2~\cite{oquab2023dinov2} (0.698 $\rightarrow$ 0.704) and CLIP-Aesthetic~\cite{clipaes} (5.665 $\rightarrow$ 5.887) metrics, with an observed latency of 159 seconds.


\begin{figure*}[!t]
    \centering
    \includegraphics[width=\linewidth]{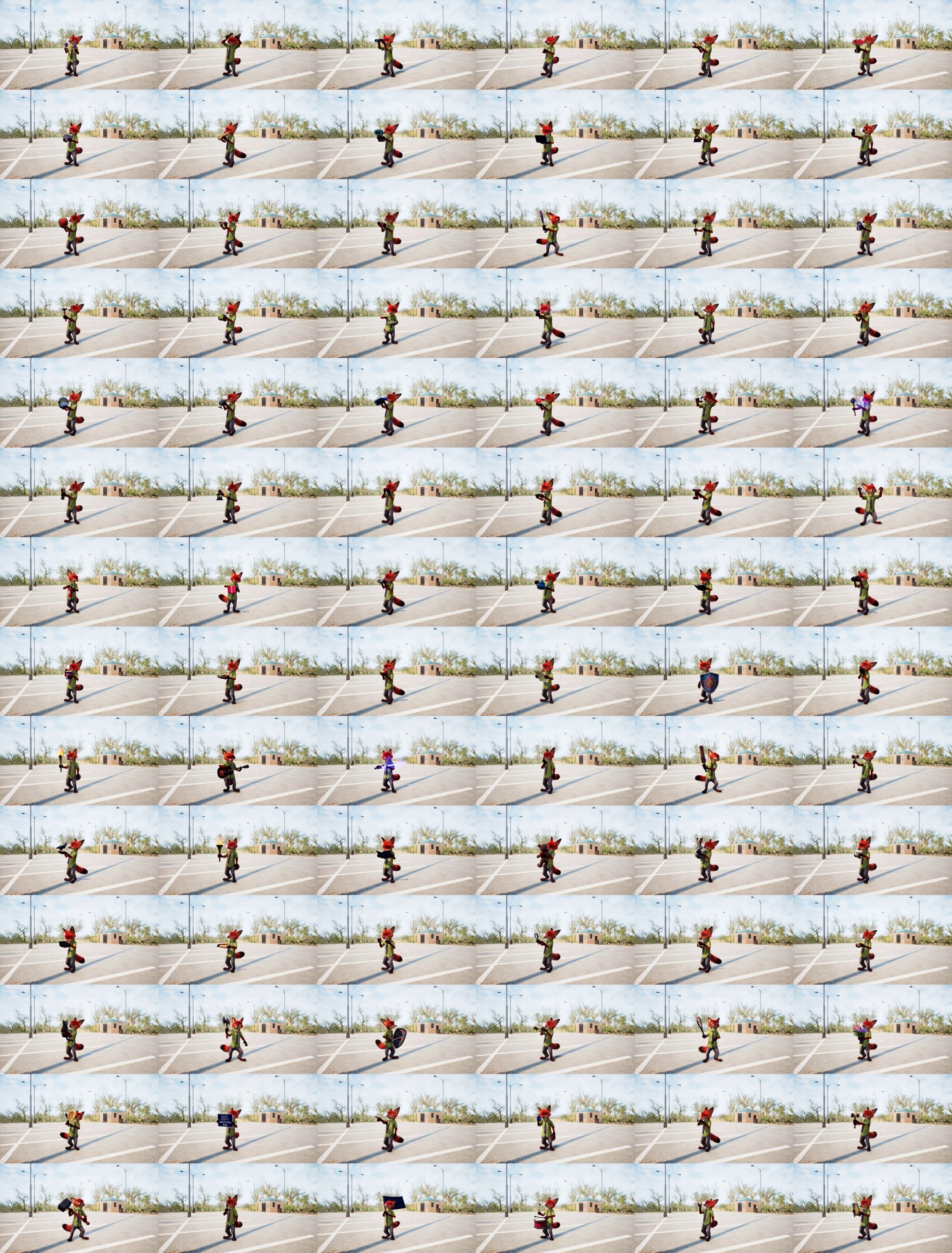}
    \caption{Visualization of a character performing 84 randomly selected novel actions.}
    \label{fig:vis_ood_action}
\end{figure*}

\subsection{Overall Inference Latencies}
While the main paper and the appendix \cref{sec:inf_acc_supp} focus on the inference overhead of CustomX, this section provides a comprehensive temporal analysis under cold-start conditions—specifically, scenarios where 3DGS scene assets and scene video rendering have not been pre-computed. Evaluated on a single NVIDIA H100 GPU, the total latency for synthesizing a 96-frame video at 360P resolution is partitioned into: (1) an initial one-time 3DGS scene generation phase (approximately 300 seconds via World Labs Marble~\cite{worldlabs} or roughly 9 seconds via FlashWorld~\cite{li2025flashworld}); (2) 3DGS rendering (approximately 0.695 seconds); (3) the accelerated model inference (approximately 21 seconds). We emphasize that: (a) Step (1) can be bypassed if pre-existing 3DGS assets are utilized; (b) Step (1) represents a one-time computational cost, whereas Steps (2) and (3) are executed iteratively; and (c) in the absence of our proposed acceleration framework, the identical 96-frame generation requires approximately 121 seconds.

\subsection{Computational Cost of 720P Video Generation}
Owing to the substantial token count, generating high-resolution (720P) videos remains computationally demanding for current open-source video generation foundation models. Specifically, Transformer-based architectures exhibit significant inference latency when processing long-sequence spatiotemporal data. For instance, synthesizing a 96-frame video at 720P resolution requires the processing of approximately 85,800 spatiotemporal tokens; this incurs a temporal overhead of roughly 151 seconds for HunyuanCustom-13B~\cite{hu2025hunyuancustom} and 182 seconds for Wan2.2-14B~\cite{wan2025} on 8× NVIDIA H100 GPUs. We acknowledge this limitation for both existing video generators and our method. As detailed in the main paper, we provide models trained at both 360P and 720P resolutions. The 360P variant serves as the default configuration—facilitated by DMD2~\cite{yin2024improved} for accelerated inference—whereas the 720P model is reserved for scenarios requiring enhanced visual quality.

\section{More Visualizations}


\subsection{Action Control and Generalization}
\begin{figure*}[!t]
    \centering
    \includegraphics[width=\linewidth]{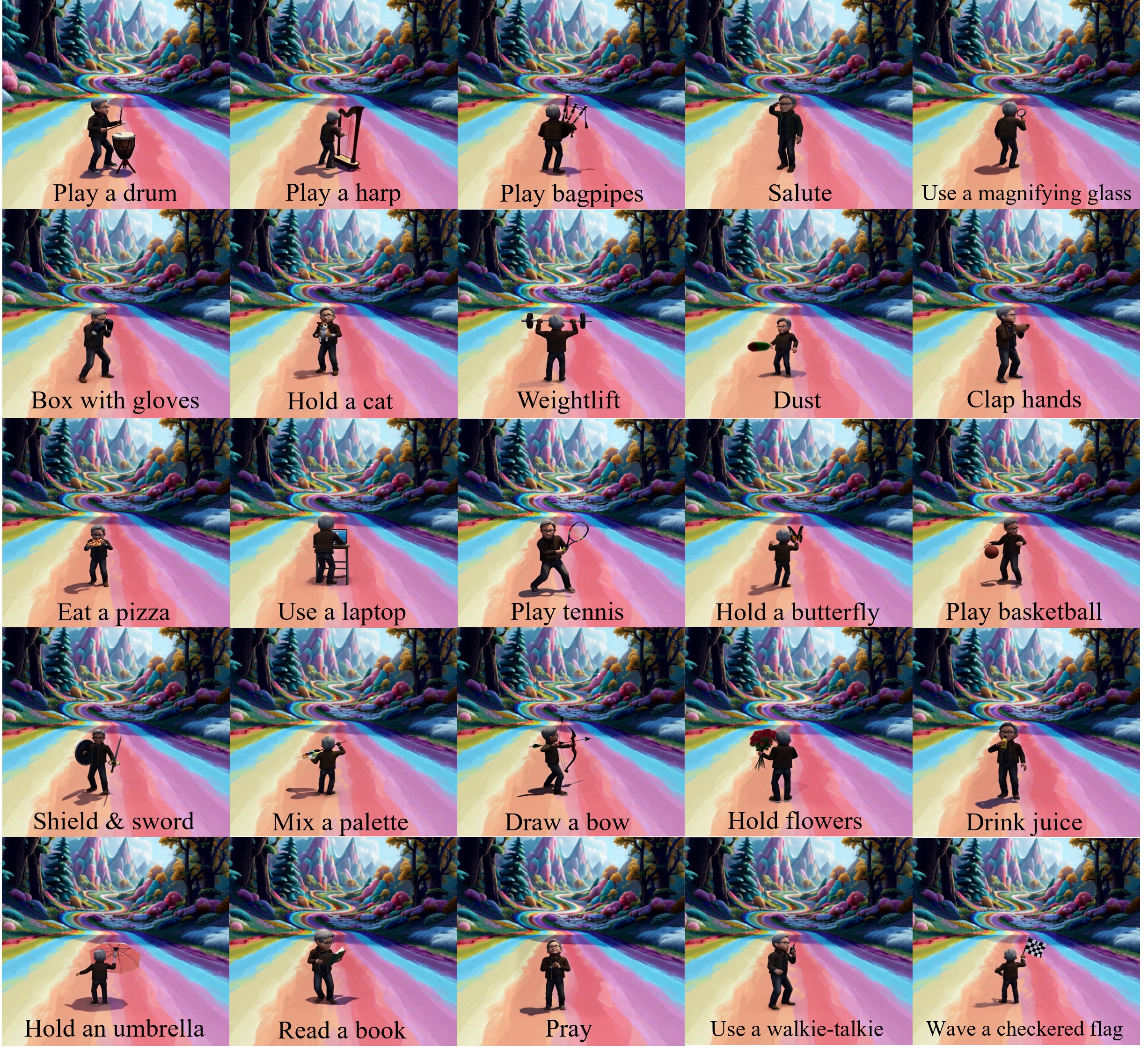}
    \caption{Visualization of a character performing 25 randomly selected novel actions with text annotations.}
    \label{fig:action_generalization}
\end{figure*}

In the main paper, we quantitatively report the success rate of controlling a character to perform 142 novel actions. In \cref{fig:vis_ood_action}, we provide qualitative results by visualizing 84 randomly selected actions using the same character. \cref{fig:action_generalization} further illustrates the model’s generalization by showing 25 randomly selected actions—with text annotations—performed by a different character.

\begin{figure*}[!t]
    \centering
    \includegraphics[width=\linewidth]{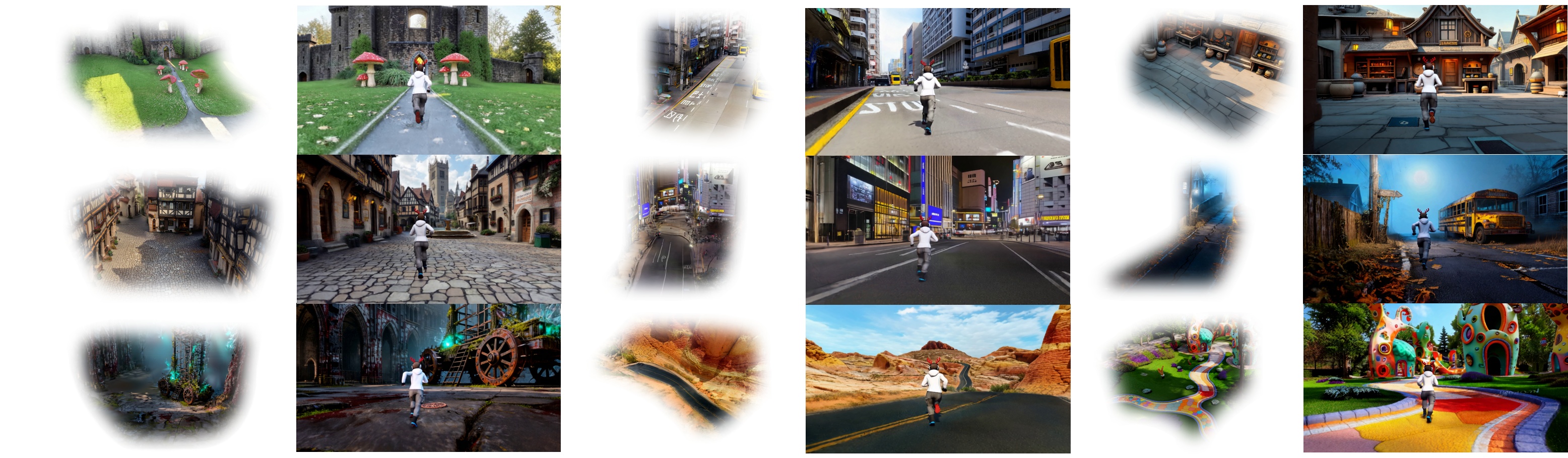}
    \caption{Visualization of a character exploring various 3DGS worlds.}
    \label{fig:scene_customization}
\end{figure*}

\begin{figure*}[!t]
    \centering
    \includegraphics[width=\linewidth]{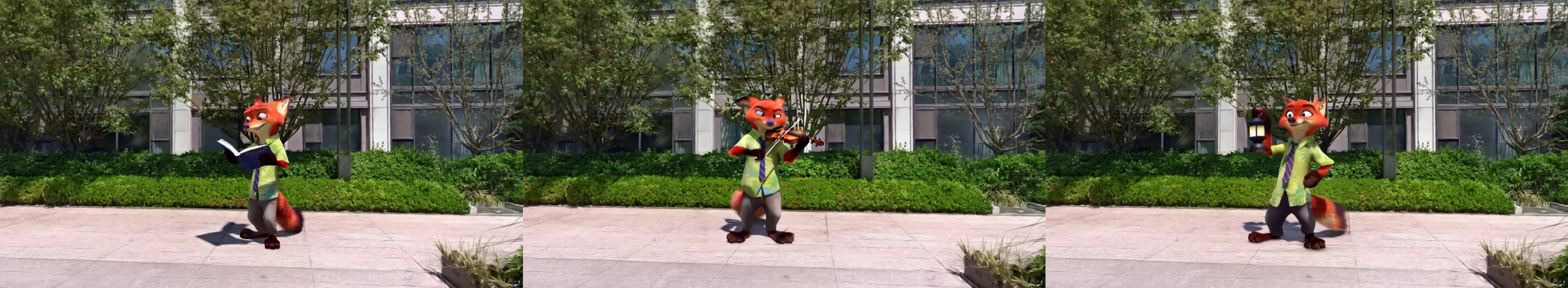}
    \caption{Visualization of a character performing novel actions in a publicly available reconstructed real-world 3DGS scene.}
    \label{fig:real_world_scene}
\end{figure*}

\subsection{Scene Customization}
Our model supports flexible scene customization. Using state-of-the-art 3DGS scene generators, users can create diverse environments and control any character to explore these worlds. In this work, most 3DGS scenes are sourced from World Labs Marble~\cite{worldlabs}. \cref{fig:scene_customization} shows examples of characters navigating a variety of generated worlds.
The proposed model also exhibits compatibility to real-world environments. \cref{fig:real_world_scene} demonstrates visualization built upon a publicly available reconstructed real-world 3DGS scene.


\subsection{Character Customization}
\begin{figure*}[!t]
    \centering
    \includegraphics[width=\linewidth]{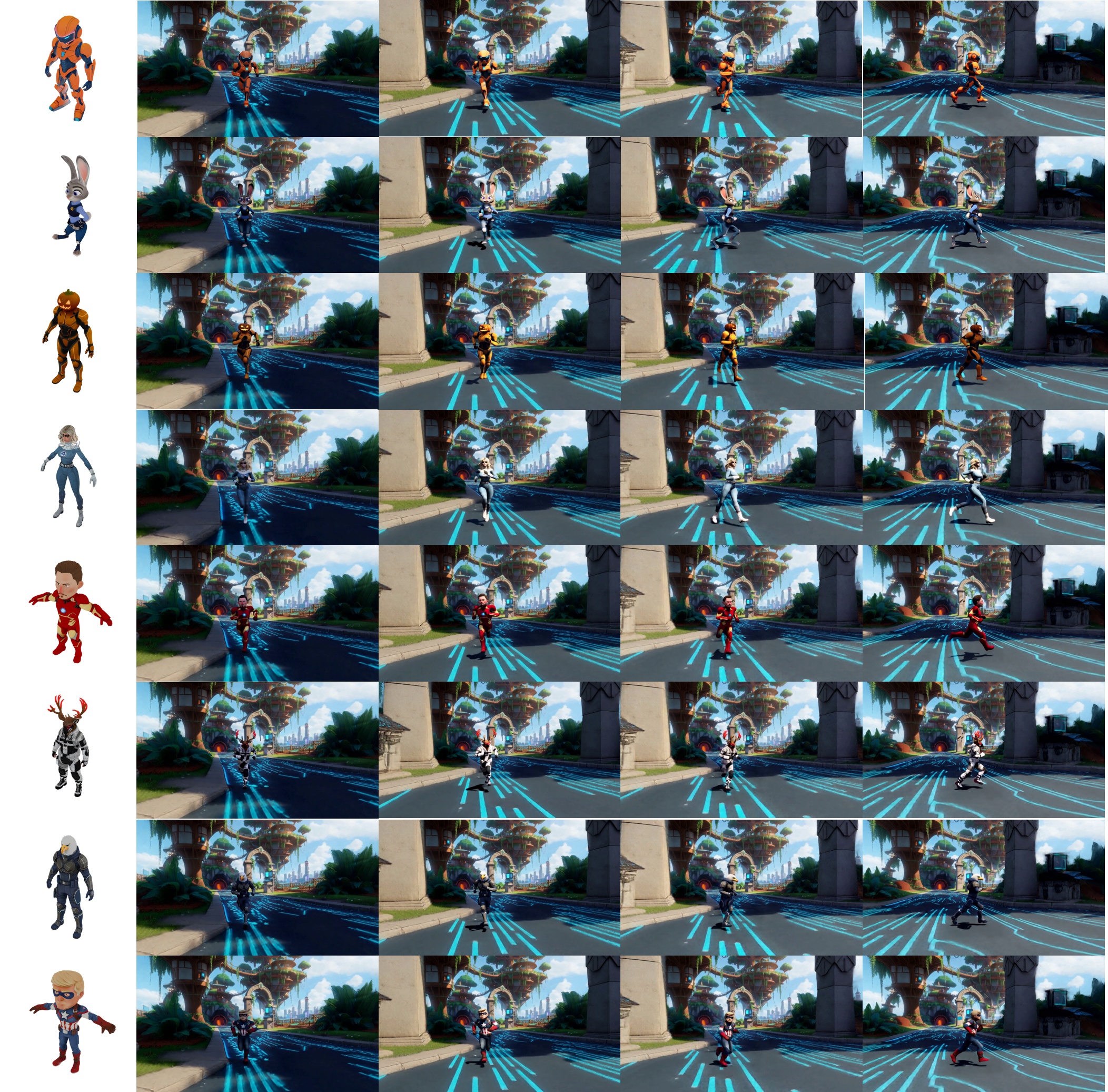}
    \caption{Visualization of diverse characters performing locomotion actions (Part 1).}
    \label{fig:character_customization_0}
\end{figure*}

\begin{figure*}[!t]
    \centering
    \includegraphics[width=\linewidth]{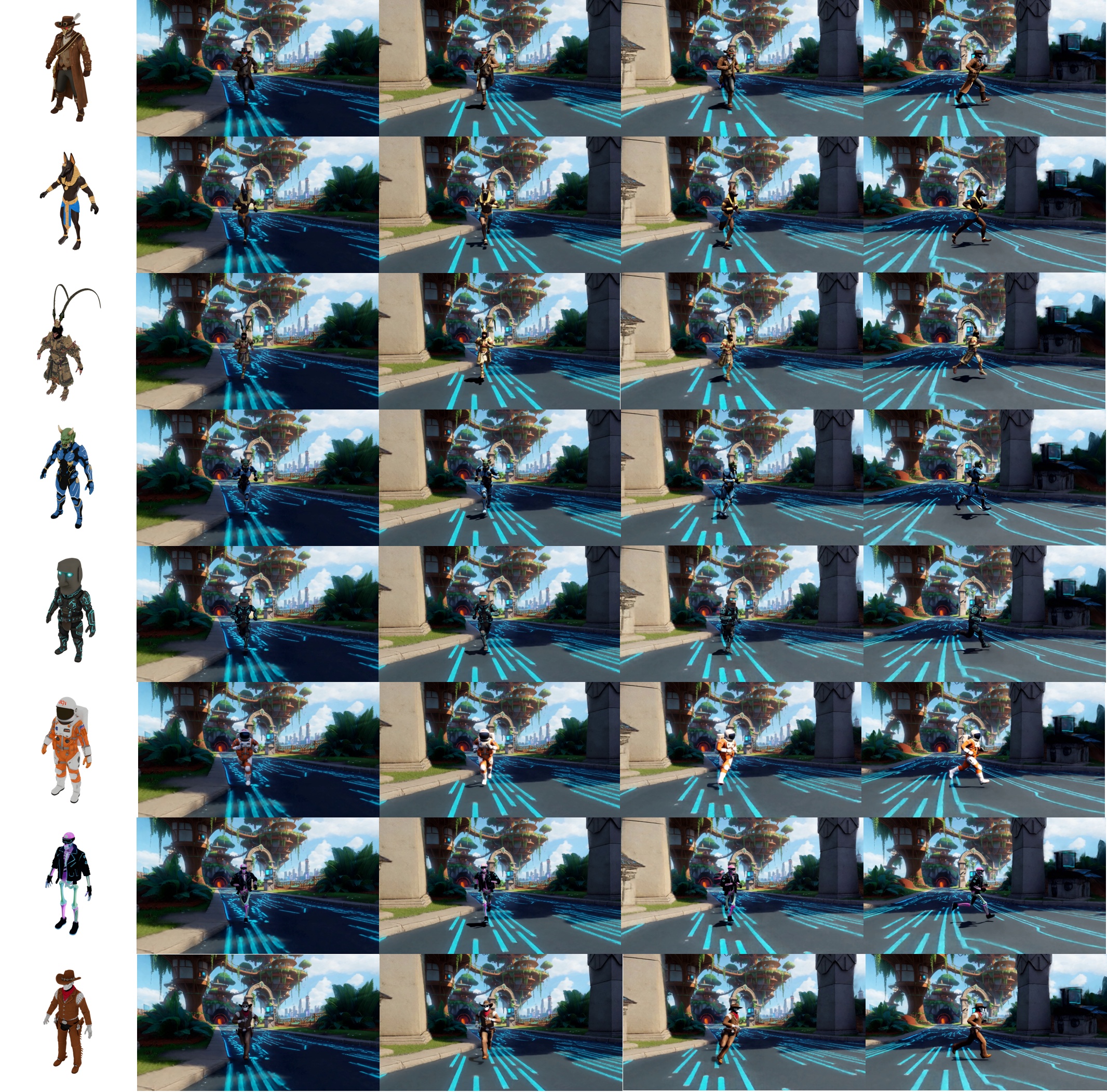}
    \caption{Visualization of diverse characters performing locomotion actions (Part 2).}
    \label{fig:character_customization_1}
\end{figure*}

Our model demonstrates strong generalization in controlling previously unseen characters. Leveraging mature 3D character-generation tools—such as Hunyuan3D \cite{hy3dweb}, Tripo~\cite{tripo}, Meshy~\cite{meshy}, and Rodin~\cite{rodin}—or sourcing assets from online repositories like Sketchfab~\cite{sketchfab}, diverse 3D characters can be easily acquired and used directly for inference. \cref{fig:character_customization_0} and \cref{fig:character_customization_1} illustrate examples of these characters performing locomotion actions.


\subsection{Long-Horizon Generation}
\begin{figure*}[!t]
    \centering
    \includegraphics[width=\linewidth]{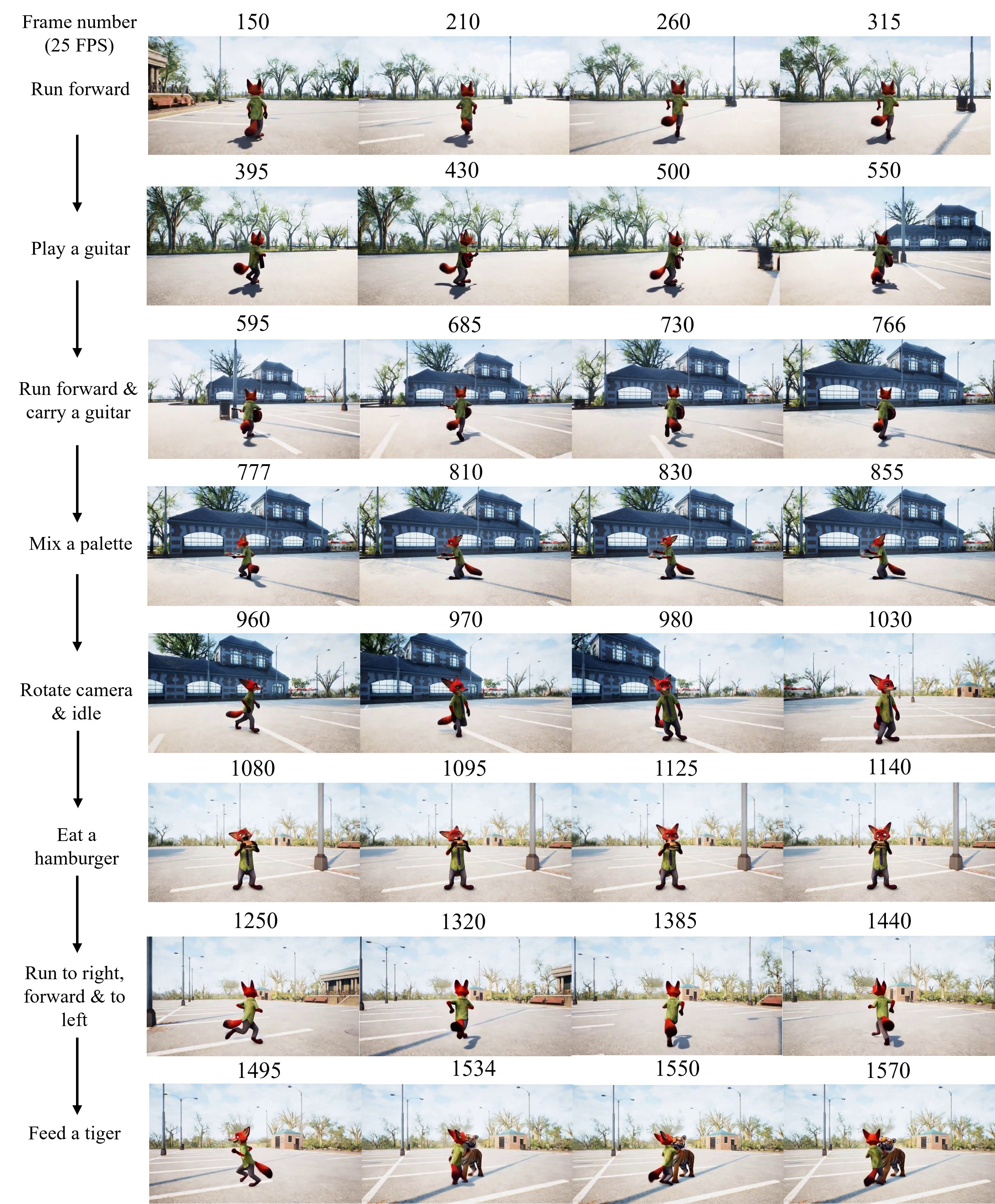}
    \caption{Visualization of long-horizon generation (Example 1).}
    \label{fig:long_gen_0}
\end{figure*}

\begin{figure*}[!t]
    \centering
    \includegraphics[width=\linewidth]{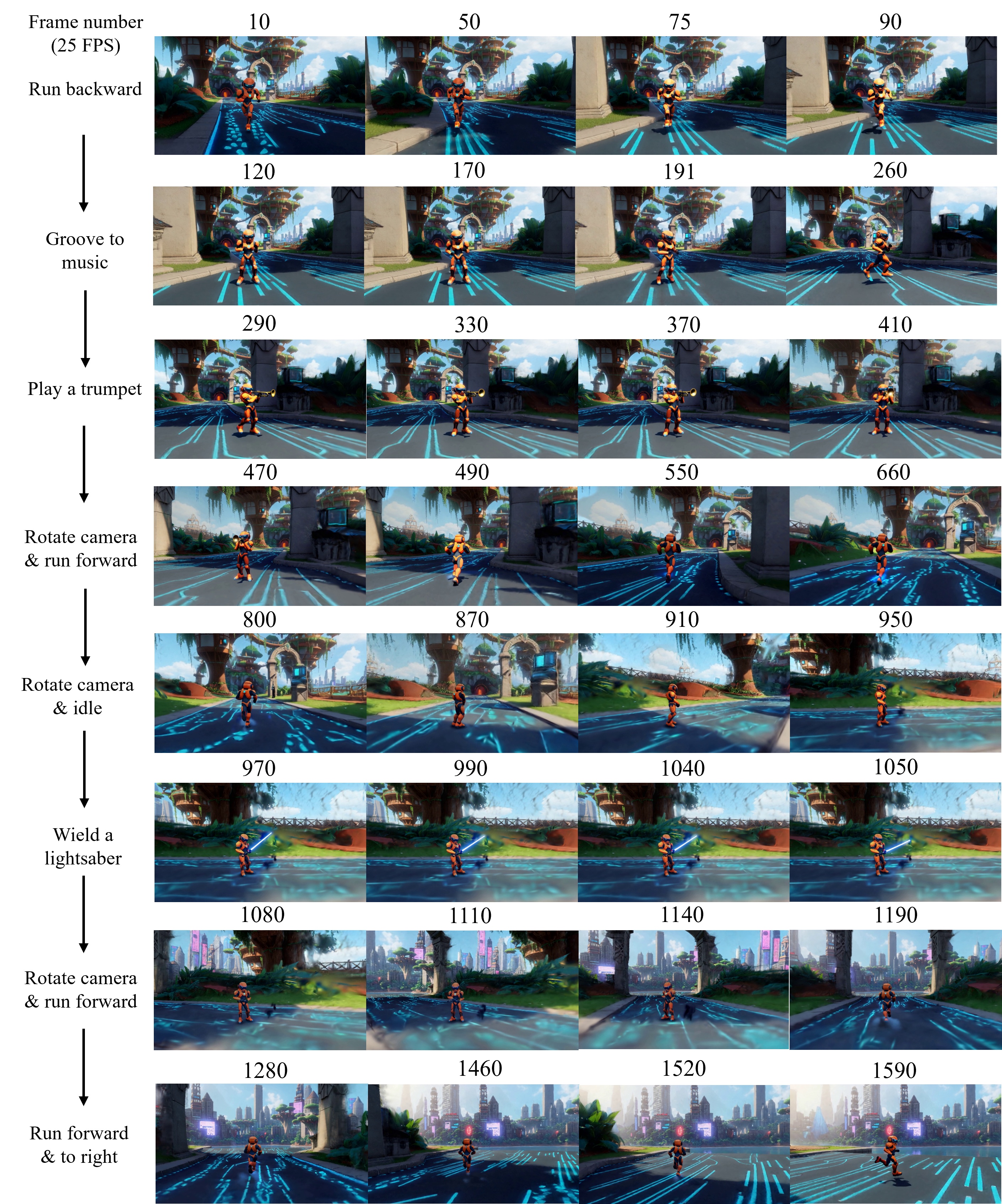}
    \caption{Visualization of long-horizon generation (Example 2).}
    \label{fig:long_gen_1}
\end{figure*}

Our model supports auto-regressive generation, enabling the creation of temporally coherent video sequences that build upon previously generated clips. This capability allows for extended, long-horizon user–model interactions. \cref{fig:long_gen_0} and \cref{fig:long_gen_1} present two examples of long-horizon generation.

\section{Potential Negative Societal Impact}
In line with existing video generation models, our framework may be repurposed to produce malicious content, highlighting the importance of ethical safeguards.

%
%
\bibliographystyle{splncs04}
\bibliography{main}
\end{document}